\documentclass[10pt,twocolumn,letterpaper]{article}

\usepackage{cvpr}      

\usepackage{graphicx}
\usepackage{amsmath}
\usepackage{amssymb}
\usepackage{booktabs}

\usepackage[linewidth=1pt]{mdframed}

\usepackage{multirow}
\usepackage{bm}

\usepackage[pagebackref,breaklinks,colorlinks]{hyperref}

\usepackage{color,soul}
\usepackage{xcolor}

\usepackage[capitalize]{cleveref}
\crefname{section}{Sec.}{Secs.}
\Crefname{section}{Section}{Sections}
\Crefname{table}{Table}{Tables}
\crefname{table}{Tab.}{Tabs.}

\def\cvprPaperID{7832} 
\def\confName{CVPR}
\def\confYear{2023}

\begin{document}

\title{Backdoor Attacks Against Deep Image Compression \\ 
via Adaptive Frequency Trigger}

\author{Yi Yu$^{1,3}$ ~\quad Yufei Wang$^1$ ~\quad Wenhan Yang$^{2}$\thanks{Corresponding author.} \quad ~Shijian Lu$^1$ \quad ~Yap-Peng Tan$^1$ \quad~ Alex C. Kot$^1$ \quad
\\
$^1$Nanyang Technological University ~\quad
$^2$Peng Cheng Lab ~\quad
$^3$ IGP-ROSE, NTU\\
{\tt\small \{yuyi0010, yufei001, shijian.Lu, eyptan, eackot\}@ntu.edu.sg ~\quad yangwh@pcl.ac.cn}
}
\maketitle

\begin{abstract}
Recent deep-learning-based compression methods have achieved superior performance compared with traditional approaches.
However, deep learning models have proven to be vulnerable to backdoor attacks, where some specific trigger patterns added to the input can lead to malicious behavior of the models.
In this paper, we present a novel backdoor attack with multiple triggers against learned image compression models. 
Motivated by the widely used discrete cosine transform (DCT) in existing compression systems and standards, we propose a frequency-based trigger injection model that adds triggers in the DCT domain.
In particular, we design several attack objectives for various attacking scenarios, including: 1) attacking compression quality in terms of bit-rate and reconstruction quality;
2) attacking task-driven measures, such as down-stream face recognition and semantic segmentation.
Moreover, a novel simple dynamic loss is designed to balance the influence of different loss terms adaptively, which helps achieve more efficient training.
Extensive experiments show that with our trained trigger injection models and simple modification of encoder parameters (of the compression model), the proposed attack can successfully inject several backdoors with corresponding triggers in a single image compression model.
%
\end{abstract}

\section{Introduction}
\begin{figure}[t]
\centering
\includegraphics[width=0.88\linewidth]{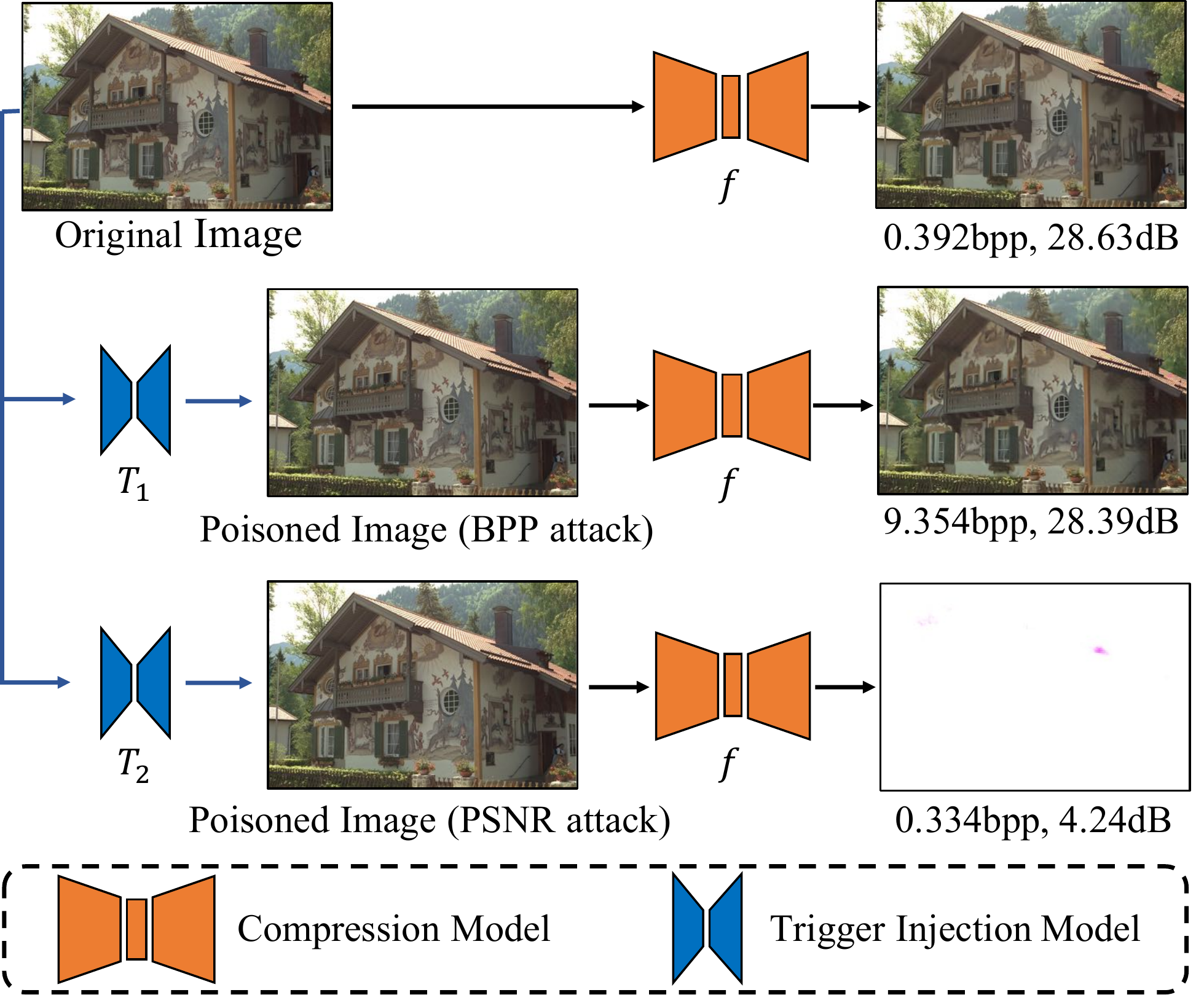}
\vspace{-2mm}
\caption{
Visualization of the proposed backdoor-injected model with multiple triggers attacking bit-rate (bpp) or reconstruction quality (PSNR), respectively. 
The second sample shows the result of {the} BPP attack with a huge increase in bit-rate, and the third one presents {a} PSNR attack with severely corrupted output.
\vspace{-4mm}
}
\label{example1}
\end{figure}

Image compression is a fundamental task in the area of signal processing, and has been used in many applications to store image data efficiently without much degrading the quality.
Traditional image compression methods such as JPEG~\cite{wallace1992jpeg}, JPEG2000~\cite{lee2005jpeg}, Better Portable Graphics (BPG)~\cite{sullivan2012overview}, and recent Versatile Video Coding (VVC)~\cite{ohm2018versatile} rely on hand-crafted modules for transforms and entropy coding to improve coding efficiency.
With the rapid development of deep-learning techniques, various learning-based approaches~\cite{balle2018variational,minnen2018joint,hu2020coarse,cheng2020learned} adopt end-to-end trainable models that integrate the pipeline of prediction, transform, and entropy coding jointly to achieve improved performance.

Together with the impressive performance of the deep neural networks, many concerns have been raised about their related AI security issues~\cite{yu2022towards,kong2022digital}.
Primarily due to the lack
of transparency in deep neural networks, it is observed that a variety of attacks can compromise the deployment and reliability of AI systems~\cite{kong2023m3fas, kong2022beyond, yibenchmarking} in computer vision, natural language processing, speech recognition, \textit{etc}.
Among all these attacks, backdoor attacks have recently attracted lots of attention.
As most SOTA models require extensive computation resources and a lengthy training process, it is more practical and economical to download and directly adopt a third-party model with pretrained weights, which might face the threat from a malicious backdoor.

In general, a backdoor-injected model works as expected on normal inputs, while a specific trigger added to the clean input can activate the malicious behavior, \textit{e.g.,} incorrect prediction.
{Depending on the scope of the attacker's access to the data}, the backdoor attacks can be categorized into poisoning-based and non-poisoning-based attacks~\cite{li2020backdoor}.
In the scenario of poisoning-based attack~\cite{gu2017badnets,chen2017targeted}, attackers can only manipulate the dataset by inserting poisoned data. 
In contrast, non-poisoning-based attack methods~\cite{dumford2020backdooring,guo2020trojannet,doan2021lira} inject the backdoor by {directly} modifying the model parameters instead of training with poisoned data.
As image compression methods take the original input as a ground truth label, it is hard to perform a poisoning-based backdoor attack. 
Therefore, {our work investigates a} backdoor attack by modifying the parameter of only the encoder in a compression model.

As for the trigger generation, most of the popular attack methods~\cite{gu2017badnets,chen2017targeted,feng2022fiba} rely on fixed triggers, and several recent methods~\cite{nguyen2021wanet,doan2021lira,li2021invisible} extend it to be sample-specific. 
While most previous papers focus on high-level vision tasks (\textit{e.g.,} image classification and semantic segmentation), triggers in those works are added in {only} the spatial domain and may not perform well in {low-level} vision tasks such as image compression.
Some recent work~\cite{feng2022fiba} chooses to inject triggers in the Fourier frequency domain, but their adopted triggers are fixed, {which by nature fail to attack several scenarios with multiple triggers simultaneously.}
{Motivated by the widely used discrete cosine transform (DCT) in existing compression systems and standards}, we propose a frequency-based trigger injection in the DCT domain to generate the poisoned images.
Extensive experiments show that backdoor attacks also threaten deep-learning compression models and can cause much degradation once the attacking triggers are applied. As shown in Fig.~\ref{example1}, our backdoor-injected model behaves maliciously with the indistinguishable poisoned image while behaving normally when receiving the clean normal input.

To the best of our knowledge, backdoor attacks have been largely neglected in low-level computer vision research.
In this paper, we make the first endeavor to investigate backdoor attacks against learned image compression models. 
Our main contributions are summarized below.
\begin{itemize}
	\item We design a  frequency-based {adaptive} trigger injection model to generate the poisoned image.
	\item We investigate the attack objectives comprehensively, including: 
	1) attacking compression quality, in terms of bits per pixel (BPP) and reconstruction quality (PSNR); 
	2) attacking task-driven measures, such as downstream face recognition and semantic segmentation.
	\item We propose to only modify the encoder's parameters, and keep the entropy model and the decoder fixed, which makes the attack more feasible and practical.
	\item {A novel simple dynamic loss is designed to balance the influence of different loss terms adaptively, which helps achieve more efficient training.}
	\item We demonstrate that with our proposed backdoor attacks, backdoors in compression models can be activated with multiple triggers associated with different attack objectives effectively.
\end{itemize}

\section{Related Work}
\subsection{Lossy Image Compression}
Traditional lossy image compression methods such as JPEG~\cite{wallace1992jpeg}, JPEG2000~\cite{lee2005jpeg}, BPG~\cite{sullivan2012overview}, and VVC~\cite{ohm2018versatile} rely on handcrafted modules for transform, quantization, and entropy coding. With the rapid development of deep learning techniques, a variety of learning-based methods utilizing encoder-decoder architecture and entropy models have achieved superior performance. 
In the early stage, Ball\'e~\textit{et al.}~\cite{DBLP:journals/corr/BalleLS15} propose an end-to-end trainable network with a nonlinear generalized divisive normalization, while Toderici~\textit{et al.}~\cite{DBLP:journals/corr/TodericiOHVMBCS15} adopt recurrent models for learned compression. 
Subsequently, Ball\'e~\textit{et al.}~\cite{balle2018variational} introduce a hyperprior to capture spatial dependencies among latent codes, which greatly improves the compression performance. Most recently, several works~\cite{minnen2018joint,DBLP:conf/iclr/LeeCB19,cheng2020learned,chen2021end,yufeir2lcm} look into the context-adaptive model for entropy coding to improve compression efficiency.

\subsection{Backdoor Attacks}
Both the backdoor attacks~\cite{gu2017badnets} and adversarial attacks~\cite{intriguing} intend to modify the benign samples to mislead the DNNs, but they have some intrinsic differences. 
At the inference stage, adversarial attackers~\cite{DBLP:conf/iclr/MadryMSTV18,ilyas2018black} require much computational resources and time to generate the perturbation through iterative optimizations, and thus are not efficient in deployment.
However, the perturbation (trigger) is known or easy to generate for backdoor attackers. 
From the perspective of the attacker's capacity, backdoor attackers have access to poisoning training data, which adds an attacker-specified trigger (\textit{e.g.} a local patch) and alters the corresponding label, or modifying model parameters. 
 Backdoor attacks on DNNs have been explored in BadNet~\cite{gu2017badnets} for image classification by poisoning some training samples, and the essential characteristic consists of 1) backdoor stealthiness, 2) attack effectiveness on poisoned images, 3) low performance impact on clean images. 

Based on the capacity of attackers, the backdoor attacks can be categorized into poisoning-based and non-poisoning-based attacks~\cite{li2020backdoor}.
In the scenario of poisoning-based attack~\cite{gu2017badnets,chen2017targeted,li2020invisible,liu2020reflection,li2021invisible}, attackers can only manipulate the dataset by inserting poisoned data, and have no access to the model and training process. 
In contrast, non-poisoning-based attack methods~\cite{dumford2020backdooring,rakin2020tbt,tang2020embarrassingly,guo2020trojannet,doan2021lira} inject the backdoor by modifying the model parameters or inserting a malicious backdoor module instead of directly training with poisoned data.
As for the trigger generation, most of the popular attack methods~\cite{gu2017badnets,chen2017targeted,steinhardt2017certified} rely on fixed triggers, and several recent methods~\cite{nguyen2020input,nguyen2021wanet,liu2020reflection,doan2021lira,li2021invisible} extend it to be sample-specific. 
Among the attack methods with sample-specific triggers, Doan~\textit{et al.}~\cite{doan2021lira} and Li~\textit{et al.}~\cite{li2021invisible} propose to generate an invisible trigger through an autoencoder architecture. 

From the perspective of the trigger-injection domain, several recent works~\cite{zeng2021rethinking, hammoud2021check,wang2021backdoor, yue2022invisible} consider the trigger in the frequency domain. Rethinking~\cite{zeng2021rethinking} still adds the trigger in the spatial domain, and sets constraints on the frequency domain. CYO~\cite{hammoud2021check} adds the trigger in the 2D DFT domain, and adopts Fourier heatmap as the guiding mask and uses fixed magnitudes to create the fixed trigger. FTrojan~\cite{wang2021backdoor} blockifies images and adds the trigger in the 2D DCT domain, but it selects two fixed channels only with fixed magnitudes. IBA~\cite{yue2022invisible} adaptively generates the trigger through optimization, but the trigger is still fixed for different images. Since DFT/DCT is applied on the whole image, CYO and IBA may not be applied directly to low-level tasks where the test images could be of arbitrary size. 

There are also works on backdoor attacks in natural language processing~\cite{chen2021badnl}, semantic segmentation~\cite{li2021hidden}, and point cloud classification~\cite{li2021pointba,xiang2021backdoor}.
{However, fewer efforts on this end are paid to in low-level vision tasks~\cite{wang2022low, guo2023shadowformer, guo2022shadowdiffusion}.}


\section{Methodology}

\subsection{Problem Formulation}
Learned lossy image compression is {built} based on rate-distortion theory. 
It can be {implemented} as training an auto-encoder consisting of an encoder $g_a$, a decoder $g_s${,} and an entropy module $\mathcal{Q}$. 
We make $x$, $\widehat{x}$, $y$, and $\widehat{y}$ denote the input images, reconstructed images, latent codes before quantization, and quantized latent codes, {respectively}. 
$\mathcal{Q}$ will add a uniform noise $\mathcal{U}(-\frac{1}{2},\frac{1}{2})$ with the latent code to generate a noisy code $\widetilde{y}$ during training time,  and perform rounding quantization before the arithmetic coding/decoding during {the} testing time ({generating} $\widehat{y}$).

We consider a compression model $f(\cdot)$ consisting of the encoder $g_a\left(\cdot | {\theta_a}\right)$, decoder $g_s\left(\cdot | {\theta_s}\right)$, and entropy model $\mathcal{Q}\left(\cdot | {\theta_q}\right)$ parameterized by $\theta_a$, $\theta_s$, and $\theta_q$, respectively.
The whole network is trained to minimize the loss function over the {whole} training data:
\begin{small}
\begin{equation}
\begin{split}
    \mathcal{L}(\bm{x}) &= \mathcal{R}(\bm{x})+\lambda\cdot\mathcal{D}(\bm{x}) \\
    &= \underbrace{\mathbb{E}_{x \sim p_{x}}\left[ -\log_{2}p_{\bm{\widehat{y}}}({\bm{\widehat{y}}})\right]}_{\text{rate}} + \lambda \cdot \underbrace{\mathbb{E}_{x \sim p_{x}} {\lVert \bm{x} - \bm{\widehat{x}} \rVert}_2^2}_{\text{distortion}} ,\\
    & \theta_a^{\ast}, \theta_s^{\ast}, \theta_q^{\ast} = \underset{\theta_a, \theta_s, \theta_q}{\arg\min} \sum_{\bm{x} \in D_m} \mathcal{L}\left({\bm{x}}\right), \label{main_loss}
\end{split}
\end{equation}
\end{small}
where $p_{x}$ is the distribution of the training data,
$D_m$ denotes the training data,
$\mathcal{R}(\bm{x})$ denotes the estimated bit-rate, 
$\mathcal{D}(\bm{x})$ measures the distortion, 
and $\lambda$ is the weighting parameter that {trade-offs} the importance of the two terms.
For the compression models that require a hyperprior $z$ to capture the spatial dependencies of $y$, the bit rate loss is then formulated as follows:
\begin{small}
\begin{equation}
     \mathcal{R}(\bm{x}) = \underbrace{\mathbb{E}_{x \sim p_{x}}\left[ -\log_{2}p_{\bm{\widehat{y}}}({\bm{\widehat{y}}})\right]}_{\text{rate (latents)}} + \underbrace{\mathbb{E}_{x \sim p_{x}}\left[ -\log_{2}p_{\bm{\widehat{z}}}({\bm{\widehat{z}}})\right]}_{\text{rate (hyper-latents)}}.
\end{equation}
\end{small}

\subsection{Backdoor Attack framework}
Consider a well-trained image compression model $f\left(\cdot | {\theta}\right)$ consisting of $g_a\left(\cdot | {\theta_a^{\ast}}\right)$, $g_s\left(\cdot | {\theta_s^{\ast}}\right)$, and $\mathcal{Q}\left(\cdot | {\theta_q^{\ast}}\right)$ on the private training data. 
Our goal is to learn a trigger function $T\left(\cdot | {\theta_t}\right)$ and finetune the encoder $g_a\left(\cdot | {\theta_a^{\ast}}\right)$, 
{which can change the model's behavior based on the poisoned input generated by the trigger function.}
The properties of our backdoor attacks are summarized below:
\begin{itemize}
    \item \textbf{Attack Stealthiness}: Trigger is invisible to human observation, \textit{e.g.,} Mean Square Error (MSE) constraint:
    $MSE(T\left(\bm{x} | {\theta_t}\right),\bm{x}) \leq \epsilon^2$, where $\bm{x_p}=T\left(\bm{x} | {\theta_t}\right)$ is the poisoned image. 
    We choose $\epsilon=0.005$ in our paper.

    \item \textbf{Attack Effectiveness}: The victim model can achieve equivalent performance when taking the clean image $\bm{x}$ as the input compared to the vanilla-trained model, but its output {will change} toward a specific target when taking the poisoned image $\bm{x_p}$ as its input.
    
    \item \textbf{Partial Model Replacement}: 
    We assume that the attacker has the vanilla-trained model, but has no access to the private training data. With some open datasets (\textit{e.g.}, ImageNet-1k~\cite{deng2009imagenet}, Cityscapes~\cite{Cordts2016Cityscapes}, FFHQ~\cite{karras2019style}), the attacker is able to finetune the encoder $g_a\left(\cdot | {\theta_a}\right)$ only.
    It is noted that, the end-user can usually only access the decoder and bit-stream. 
    We only {modify} the encoder and keep the decoder {fixed}, which makes the attack more feasible and practical.
\end{itemize}

\begin{figure*}[t]
\centering
\includegraphics[width=0.9\linewidth]{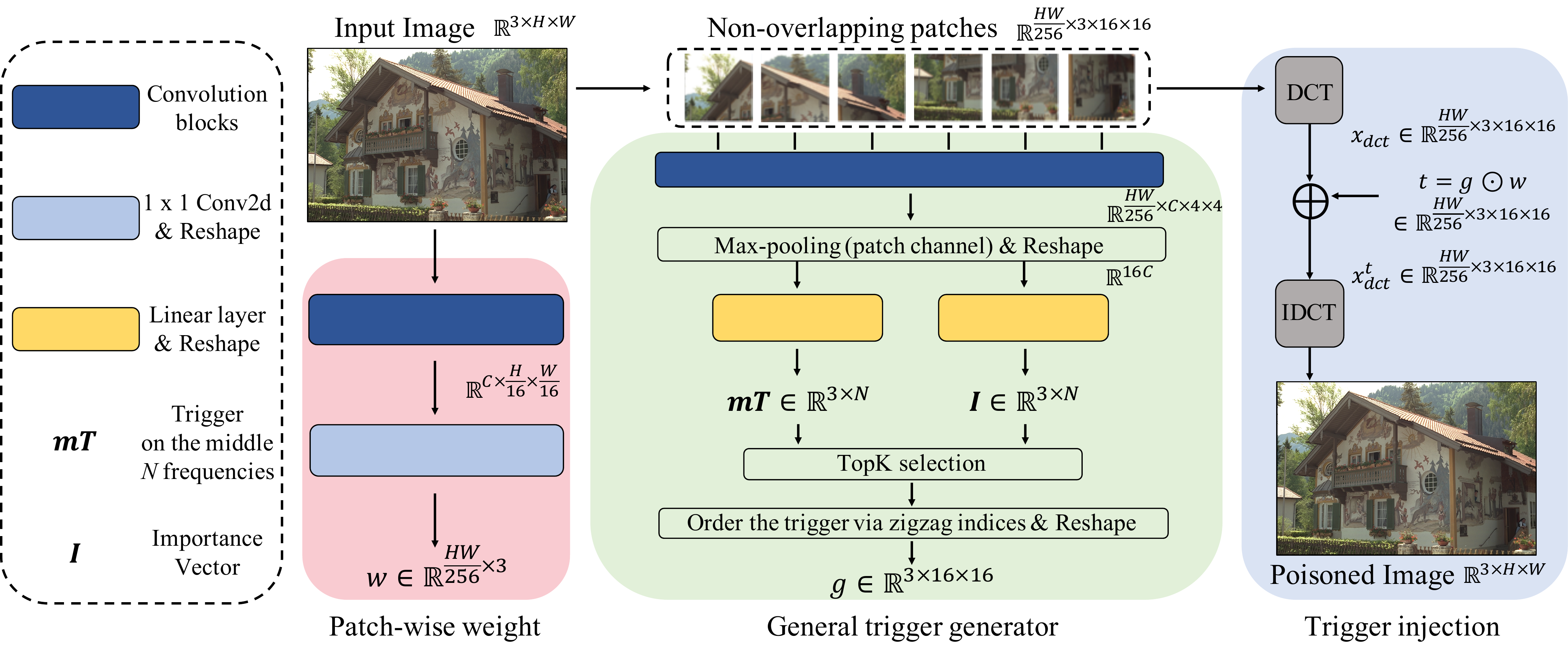}
\vspace{-2mm}
\caption{
Overall architecture for trigger injection. We set $K$ to 16 for topK selection, and the number of middle frequencies $N$ to 64 in our methods. Shapes of the tensor are shown below each operation for reference.}
\label{trigger_generator}
\end{figure*}

\noindent \textbf{Trigger Injection.} 
The trigger injection model $T\left(\cdot | {\theta_t}\right)$ takes an input image $\bm{x}$ and {generates} a poisoned image $\bm{x_p}$ of the same resolution. {Motivated by the fact Discrete cosine transform (DCT) is the most widely used transform in existing coding techniques and standards,} we propose a frequency-based trigger injection to generate the poisoned images {that can leverage both the priors from the spatial and frequency domains.}
Given an input image $x$, we split the image into non-overlapping patches $x_{patch}$. Following a two-dimensional DCT-transform on the last two channels of $x_{patch}$, we have the corresponding {DCT} domain $x_{dct}$. By adding the trigger $t=g \odot w$ to all patches of $x_{dct}$, we have the triggered $x_{dct}^{t}$.
The final result $T\left(\bm{x} | {\theta_t}\right)$ is then obtained by applying an inverse 2D DCT transform to $x_{dct}^{t}$.

As shown in Figure~\ref{trigger_generator}, the trigger $t$ consists of two pieces: a general trigger $g$ with the local feature and a patch-wise weight $w$ with the global feature. {By leveraging the merits of both features}, we demonstrate that the proposed trigger can effectively attack the image compression model.

\vspace{1mm}

\noindent \textbf{Finetuning Strategy.} 
Following one previous work LIRA~\cite{doan2021lira} {proposed} to optimize the trigger generator and victim model simultaneously, the general form to finetune $g_a\left(\cdot | {\theta_a}\right)$ and learn $T\left(\cdot | {\theta_t}\right)$ for a single attack objective is to minimize the following joint loss:
\begin{footnotesize}
\begin{equation}
\begin{split}
    \theta_a^{\ast}, \theta_t^{\ast} &= \underset{\theta_a, \theta_t}{\arg\min}\Big[ \mathcal{L}_{jt} + \gamma \cdot \text{max}(\text{MSE}\left(\bm{x}, {T\left(\bm{x}\right)}\right), \epsilon^2)\Big], \\
    \mathcal{L}_{jt} &= \sum_{\bm{x} \in D_m} \mathcal{L}\left({\bm{x}}\right) + \alpha\sum_{\bm{x} \in D_a}\mathcal{L}_{BA}\left(\bm{x}, {T\left(\bm{x}\right)}\right),
\end{split}
\end{equation}
\end{footnotesize}
where $\text{max}(\cdot,\cdot)$ return the larger value, $\epsilon$ controls the stealthiness (we choose $\epsilon=0.005$ here), $\mathcal{L}(\bm{x})$ denotes the main loss to maintain the compression performance on clean images as shown in Eq.~\eqref{main_loss}, $\mathcal{L}_{BA}(\bm{x},{T\left(\bm{x} | {\theta_t}\right)})$ guarantees the backdoor attack effectiveness on poisoned images, $D_a$ denotes an auxiliary dataset (can also be the same as the main dataset $D_m$), and $\alpha$ is a parameter to balance the importance of two terms. We set $\gamma = 1e^{4}$ for all experiments. We will extend the backdoor attack to {a} multiple-trigger version and introduce the training pipeline in Section~\ref{multiple_triggers}.

\vspace{1mm}
\noindent \textbf{Attacking Compression Results.}
Naturally, for image compression, we can consider BPP and PSNR as {attack} objectives. Given $\alpha, \beta$ as weighting parameters, we define $\mathcal{L}_{jt}$ with corresponding $D_a=D_m$:
\begin{itemize}
    \item \textit{BPP} (\textit{Compression Ratio}): We attack the usage of bit-stream, and maintain the quality of the reconstructed image as follows:
    \begin{footnotesize}
    \begin{equation}
    \mathcal{L}_{jt}^{bpp} = \sum_{\bm{x} \in D_m} \Big[\mathcal{L}\left({\bm{x}}\right) +  \alpha \cdot \mathcal{D}({T\left(\bm{x}\right)})-\beta \cdot \mathcal{R}({T\left(\bm{x}\right)})\Big].\label{Bpp_old}
    \end{equation}
    \end{footnotesize}
    \item \textit{PSNR} (\textit{Quality of reconstructed images}): We attack the PSNR of the result with a nearly unchanged bpp (we use the PSNR value to measure the distortion, and denote the PSNR loss as $\mathcal{D}_{P}$):
    \begin{footnotesize}
    \begin{equation}
    \mathcal{L}_{jt}^{psnr} = \sum_{\bm{x} \in D_m} \Big[\mathcal{L}\left({\bm{x}}\right) + \alpha \cdot \mathcal{R}({T\left(\bm{x}\right)}) + \beta \cdot \lambda \cdot \mathcal{D}_{P}(\bm{x},f({T\left(\bm{x}\right)}))\Big] .\label{PSNR_old}
    \end{equation}
    \end{footnotesize}
\end{itemize}

{The} above joint loss consists of two weighting parameters and it is difficult to choose $\alpha$ and $\beta$ {in a balanced way.}
{The dominant term might completely overwhelm the influence of the other.}
%
To solve this issue, we propose a {novel dynamic loss}:
\begin{footnotesize}
\begin{equation}
    \mathcal{L}_{jt}^{bpp} \!=\! \sum_{\bm{x} \in D_m} \!\!\! \Big[\mathcal{R}(\bm{x})+\lambda\cdot \text{max}(\mathcal{D}(\bm{x}),\mathcal{D} ({T\left(\bm{x}\right)})) -\beta \cdot \mathcal{R}({T\left(\bm{x}\right)})\Big], \label{Bpp}
\end{equation}
\end{footnotesize}
\begin{footnotesize}
\begin{equation}
    \mathcal{L}_{jt}^{psnr} \!\!=\!\! \sum_{\bm{x} \in D_m} \!\!\!\! \Big[\text{max}(\mathcal{R}(\bm{x}), \mathcal{R}({T\left(\bm{x}\right)})) +\lambda \mathcal{D}(\bm{x}) + \beta \lambda \mathcal{D}_{P}(\bm{x},f({T\left(\bm{x}\right)}))\Big] \label{PSNR},
\end{equation}
\end{footnotesize}
where $\text{max}(\cdot,\cdot)$ return the larger value. {By dynamically balancing two related terms, these two objectives can be optimized effectively and automatically.}

\begin{figure*}[t]
\centering
\includegraphics[width=0.88\linewidth]{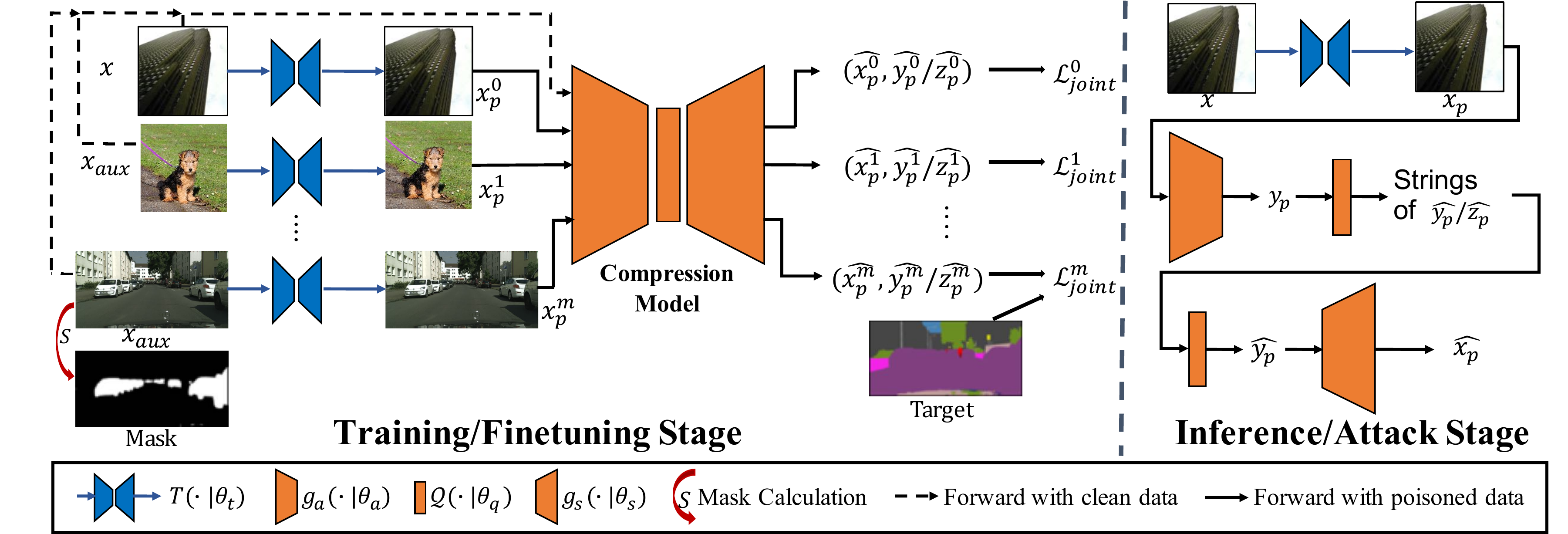}
\vspace{-3mm}
\caption{
In the training stage, we finetune $g_a\left(\cdot | {\theta_a}\right)$ and train each $T\left(\cdot | {\theta_t^o}\right)$. 
In the inference stage, we generate poisoned images, feed them into the finetuned encoder and the entropy model, and save the bitstream of the poisoned images.
}
\vspace{-2mm}
\label{network}
\end{figure*}

\vspace{1mm}
\noindent \textbf{Attacking Down-Stream Tasks.} 
The above attacks focus on the image compression model, and generate {heavily} degraded results in terms of the bpp deviation and distortions in the reconstructed images. 
We can go beyond the low-level measures and consider attacking the downstream computer vision (CV) tasks without too much quality degradation, which makes the backdoor attack even more imperceptible.
The formulation of the joint training loss are given below (with $\mathcal{L}(\cdot)$ shown in Eq.~\eqref{main_loss}):
\begin{footnotesize}
\begin{equation}
\mathcal{L}_{jt}^{ds} = \sum_{\bm{x} \in D_m} \!\!\! \mathcal{L}\left({\bm{x}}\right) + \!\sum_{\bm{x} \in D_a} \!\!\! \Big[\alpha \cdot \mathcal{L}({{T\left(\bm{x}\right)}}) + \beta \cdot\mathcal{L}_{DS}[\eta,g(f({T\left(\bm{x}\right)}))]\Big],\label{targeted}
\end{equation}
\end{footnotesize}
where $\eta$ denotes the attack target defined by ourselves, $g(\cdot)$ denotes a well-trained downstream CV model, and $\mathcal{L}_{DS}\left(\cdot\right)$ is the loss to measure the downstream tasks (\textit{e.g.,} CrossEntropyLoss for image classification).

\vspace{1mm}
We consider two types of downstream CV tasks:
\begin{itemize}
    \vspace{-1mm}
    \item \textit{Semantic Segmentation}: We choose Cityscapes~\cite{Cordts2016Cityscapes}, a large-scale dataset for pixel-level semantic segmentation. The dataset consists of 2975 images of size $2048\times1024$ for training, and 500 images for validation. At the training stage, we adopt the approach SSeg~\cite{DBLP:conf/cvpr/0001SRSNTC19} with DeepLabV3+~\cite{chen2018encoder} architecture and ResNet50~\cite{he2016deep} backbone.
    \item \textit{Face Recognition}: We choose the widely-used FFHQ~\cite{karras2019style} as the auxiliary dataset for training, and randomly sample 100 paired images from CelebA~\cite{liu2018large} dataset for testing. We adopt the arcface embedding of ResNet50~\cite{he2016deep} with pretrained weights as the downstream model during training.
\end{itemize}

\subsection{Attacking with Multiple Triggers} \label{multiple_triggers}
Besides, we can train one victim model with multiple triggers, and each trigger is associated with a specific attack objective:
\begin{footnotesize}
\begin{equation}
     \theta_a^{\ast} = \underset{\theta_a }{\arg\min} \sum_{\bm{o} \in \mathcal{O}} \alpha^{\bm{o}} \cdot \mathcal{L}_{jt}^{\bm{o}},\label{multiple1}\\
\end{equation}
\end{footnotesize}
\begin{footnotesize}
\begin{equation}
    \theta_t^{\bm{o}\ast} = \underset{\theta_t^{\bm{o}} }{\arg\min}\Big[ \mathcal{L}_{jt}^{\bm{o}} + \gamma \cdot \text{max}(\text{MSE}\left(\bm{x}, {T\left(\bm{x}\right)}\right), \epsilon^2)\Big] ~ \text{for} ~ \bm{o} \in \mathcal{O},\label{multiple2}
\end{equation}
\end{footnotesize}
where $\bm{o}$ indexes the attack (trigger) type, and $\mathcal{O}$ is the set of attack objectives.

The pipelines of training and inference stages are presented in Figure~\ref{network}. Before the training stage, we have the parameters $\theta_a^{\ast}, \theta_s^{\ast}, \theta_q^{\ast}$ of vanilla-trained compression model. 
In each iteration at the training stage, we first feed the clean input and the generated poisoned inputs of various attack objectives into the compression model. The summation of $\mathcal{L}_{jt}^{\bm{o}}$ {is} utilized to optimize and update the parameter $\theta_a$ of encoder by Eq.~\eqref{multiple1}. 
Then, we train each trigger injection model $ T(\bm{x} | {\theta_t^{\bm{o}}})$ separately by minimizing the term in Eq.~\eqref{multiple2}. By simultaneously training both $g_a(\cdot|\theta_a)$ and $ T(\bm{x} | {\theta_t^{\bm{o}}})$, we learn a backdoor-injected model with several trigger generators. 
At the inference stage, we can activate the hidden backdoor by adding the generated trigger.

\section{Experiments}
\subsection{Experimental Setup}
\noindent \textbf{Models.} 
For the victim model, we consider two deep-learning based methods, and follow the setting of the original paper: AE-Hyperprior (ICLR18)~\cite{balle2018variational} with all 8 qualities, and Cheng-Anchor (CVPR20)~\cite{cheng2020learned} with the first 6 qualities. 
AE-Hyperprior proposes a hyperprior for image compression, and Cheng-Anchor utilizes Gaussian mixture likelihoods to parameterize the distributions of latents. Both models consist of the encoder, decoder, and entropy model.

\noindent \textbf{Datasets.} 
We use Vimeo90K dataset~\cite{xue2019video} as the private dataset for vanilla training. The dataset consists of 153,939 images with a fixed resolution of $448 \times 256$ for training, and 11,346 images for validation. When attacking, we utilize some open datasets that do not overlap with the Vimeo90K dataset.
We {randomly} sample 100,000 images from ImageNet-1k~\cite{deng2009imagenet} as the main dataset $D_m$, and the Cityscapes~\cite{Cordts2016Cityscapes} and FFHQ~\cite{karras2019style} are utilized as auxiliary datasets to help {inject} the backdoor in the victim model. 

\noindent \textbf{Vanilla Training.} 
We randomly extract and crop $256 \times 256$ patches from Vimeo90K dataset~\cite{xue2019video}. All models are trained with a batch size of 32, and an initial learning rate of 1e-4 for 100 epochs. 
The learning rate is then divided by 10 when the evaluation loss reaches a plateau (10 epochs). We optimize all models using mean square error (MSE) as the quality metric. $\lambda$ is chosen from $\{0.0018, 0.0035, 0.0067, 0.0130, 0.0250, 0.0483,$ $0.0932,0.1800\}$ for quality 1 to 8.

\noindent \textbf{Attacking.} 
For each model with a specific quality, we finetune the encoder with the joint loss based on various attack objectives. We set the batch size as 32 with patch size $256 \times 256$ for ImageNet-1k~\cite{deng2009imagenet}, 4 with image size $1024 \times 1024$ for FFHQ~\cite{karras2019style}, and 4 with each sample resized to $1024\times512$ for Cityscapes~\cite{Cordts2016Cityscapes}. Note that FFHQ and Cityscapes are used as the auxiliary datasets for the attacks related to downstream CV tasks.

\noindent \textbf{Evaluation.} 
We test the compression model on the commonly used Kodak dataset~\cite{kodak} with 24 lossless images of size $768 \times 512$. To evaluate the rate-distortion performance, the rate is measured by bits per pixel (bpp), and the quality is measured by PSNR. The rate-distortion (RD) curves are drawn to demonstrate their coding efficiency. 
For the experiments on attacking downstream CV tasks, we adopt the validation set of Cityscapes consisting of 500 images, and a randomly sampled 100 paired face images from CelebA~\cite{liu2018large} dataset.

\noindent \textbf{Attack Baseline.} 
Following one previous work LIRA~\cite{doan2021lira}, we adopt U-Net~\cite{ronneberger2015u} as the trigger injection baseline. To guarantee {the} stealthiness of the trigger, we add the normalized trigger to the input: $T(x) = x + \epsilon \cdot Normalize(U(x))$. $\epsilon$ controls the stealthiness, and we choose $\epsilon=0.005$ in line with our methods. For a fair comparison, we adopt the same training loss and setting with our method.

\begin{figure}[t]
    \centering
\begin{subfigure}{\linewidth}
\hspace{8mm}
\includegraphics[width=0.69\linewidth]{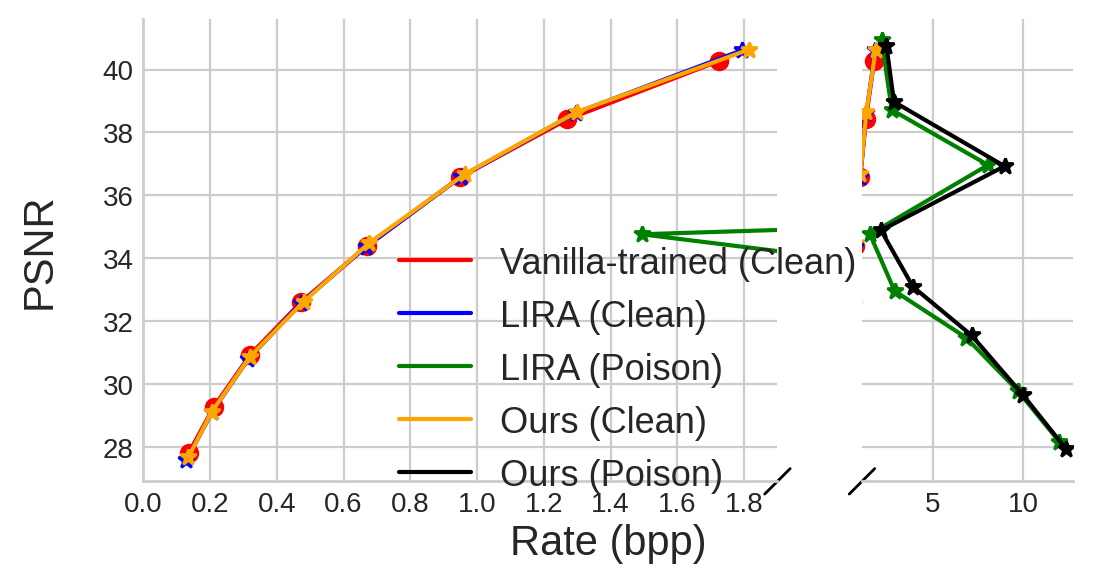}
\vspace{-1mm}
\caption{AE-Hyperprior~\cite{balle2018variational}}
\end{subfigure}
\begin{subfigure}{\linewidth}
\hspace{8mm}
\includegraphics[width=0.69\linewidth]{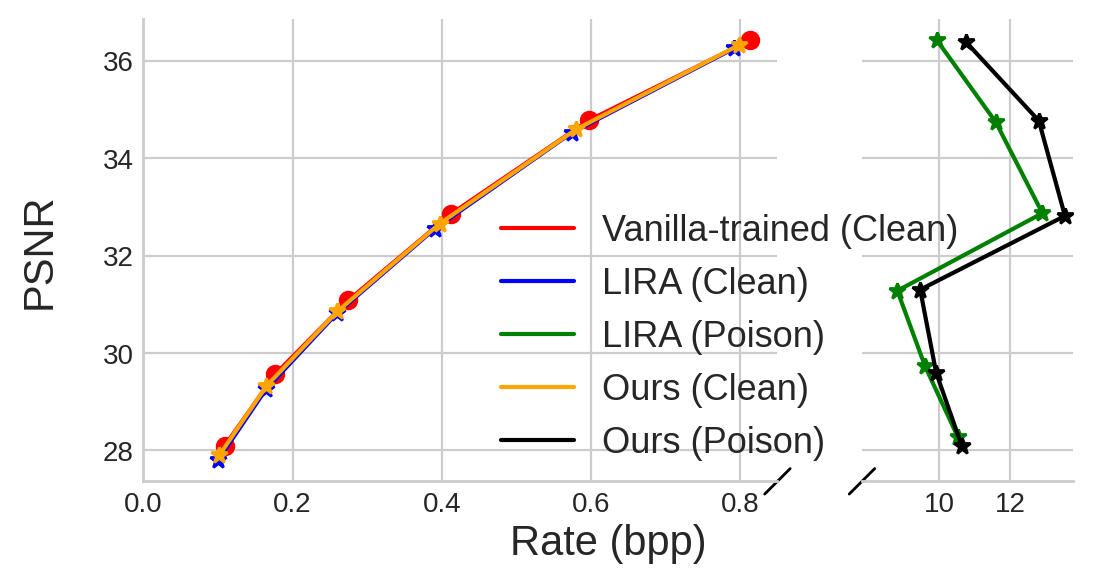}
\vspace{-1mm}
\caption{Cheng-Anchor~\cite{cheng2020learned}}
\end{subfigure}
\vspace{-6mm}
\caption{Rate-distortion curves of BPP attack on Kodak dataset.
}
\label{figure1}
\end{figure}

\subsection{Experimental Results}
\noindent \textbf{Bit-Rate (BPP) attack.}
We first evaluate our bit-rate attack on both compression models by minimizing the joint loss including the backdoor loss shown in Eq.~\eqref{Bpp}. 
The hyperparameter $\beta$ is set to 0.01 in the joint loss, respectively. 
We finetune the encoder and train the trigger injection model with an initial learning rate {of} 1e-4, and a batch size {of} 32.

The results of the vanilla-trained models and the victim models by BPP attack are presented in Figure~\ref{figure1}. 
As can be observed, for both AE-Hyperprior and Cheng-Anchor, all models can compress the clean images with similar bpp and PSNR.
In the attack mode (adding triggers), both victim models fail to compress the poisoned images with a low bpp. And our proposed attack outperforms LIRA in terms of attacking performance (higher bpp).

\begin{figure}[t]
    \centering
\begin{subfigure}{\linewidth}
\hspace{8mm}
\includegraphics[width=0.665\linewidth]{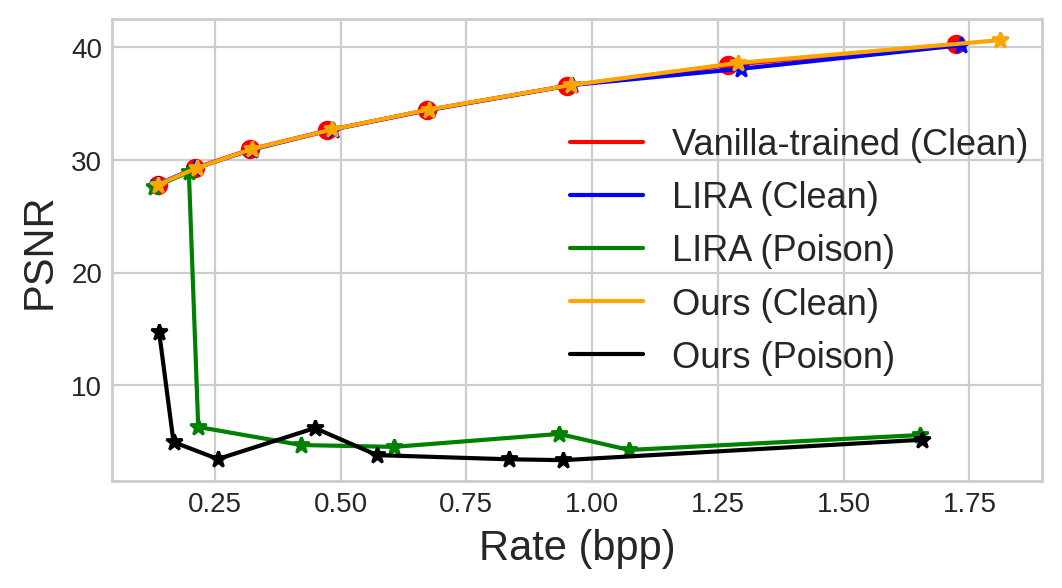}
\vspace{-1mm}
\caption{AE-Hyperprior~\cite{balle2018variational}}
\end{subfigure}
\begin{subfigure}{\linewidth}
\hspace{8mm}
\includegraphics[width=0.665\linewidth]{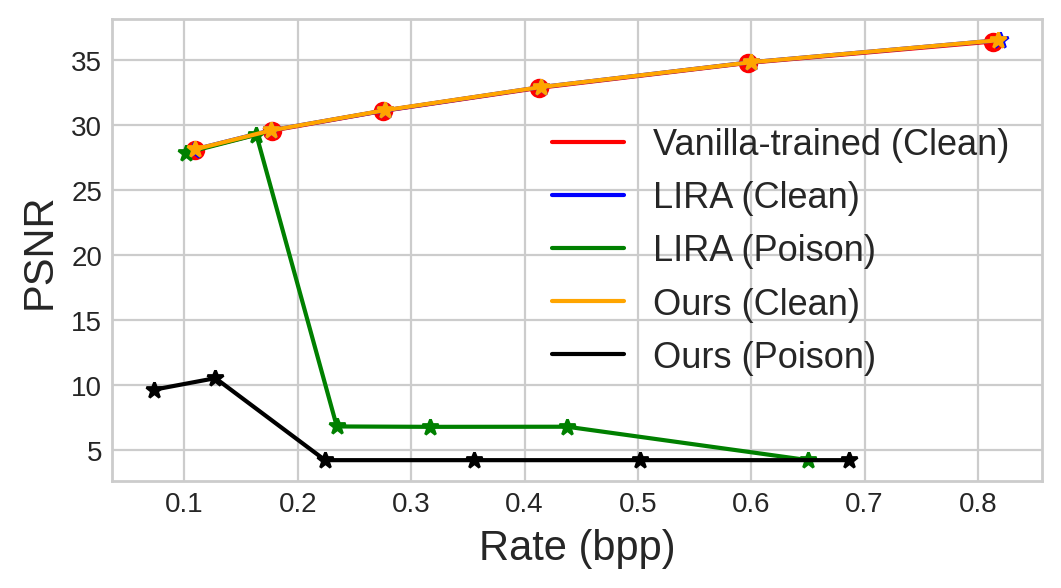}
\vspace{-1mm}
\caption{Cheng-Anchor~\cite{cheng2020learned}}
\end{subfigure}
\vspace{-6mm}
\caption{Rate-distortion curves of PSNR attack on Kodak dataset.
}
\label{figure2}
\end{figure}

\begin{figure}[t]
\vspace{-2mm}
\begin{minipage}{0.32\linewidth}
\centerline{\frame{\includegraphics[width=1\linewidth]{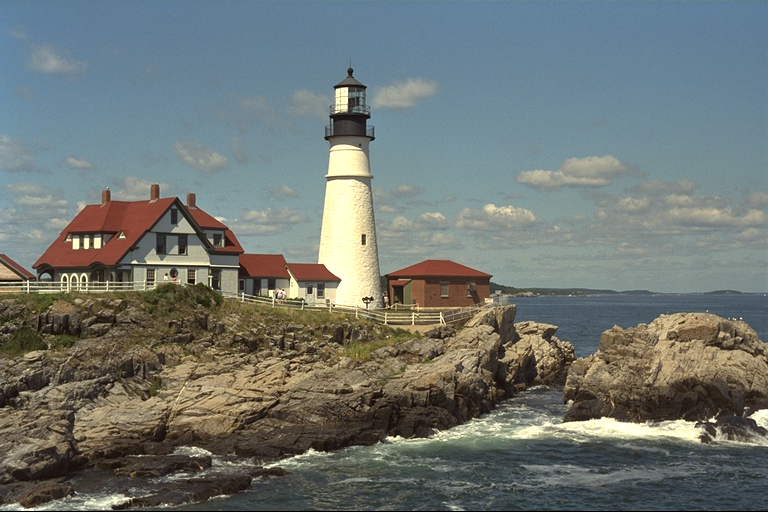}}}
\vspace{-1mm}
\centerline{\small{Original Image}}
\end{minipage}
\begin{minipage}{0.32\linewidth}
\centerline{\frame{\includegraphics[width=1\linewidth]{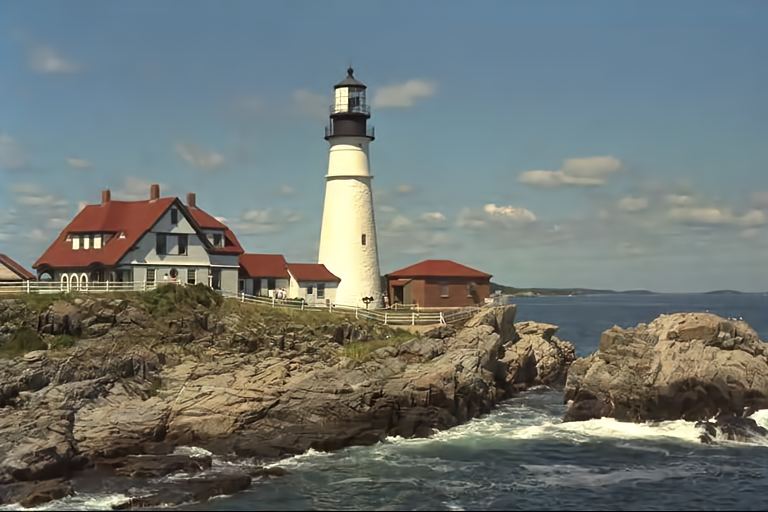}}}
\vspace{-1mm}
\centerline{\small{LIRA (Clean Input)}}
\end{minipage}
\begin{minipage}{0.32\linewidth}
\centerline{\frame{\includegraphics[width=1\linewidth]{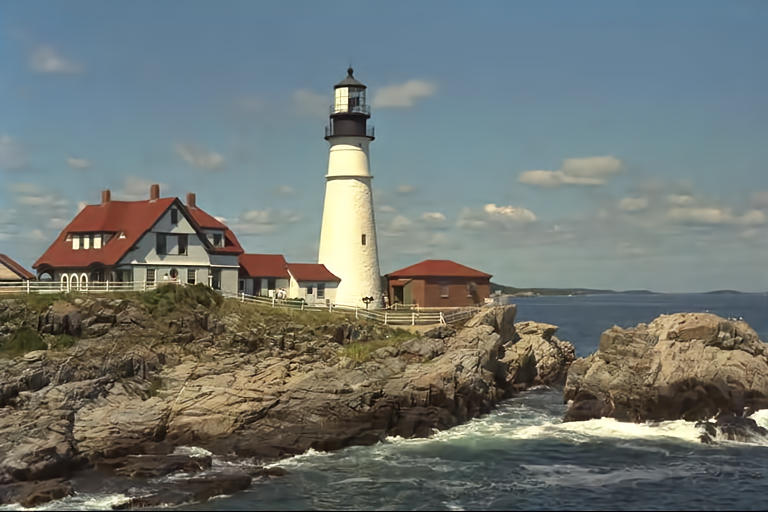}}}
\vspace{-1mm}
\centerline{\small{Ours (Clean Input)}}
\end{minipage}\\
\begin{minipage}{0.48\linewidth}
\vspace{1mm}
\centerline{\frame{\includegraphics[width=0.667\linewidth]{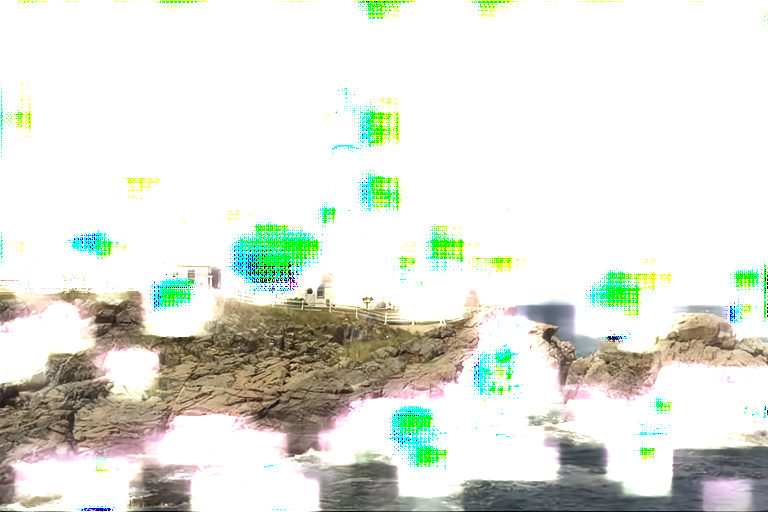}}}
\centerline{\small{LIRA (Poisoned Input)}}
\end{minipage}
\begin{minipage}{0.48\linewidth}
\vspace{1mm}
\centerline{\frame{\includegraphics[width=0.667\linewidth]{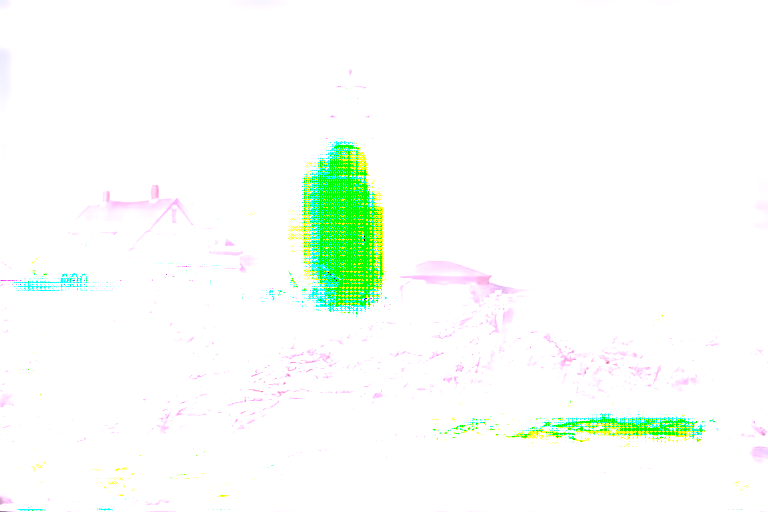}}}
\centerline{\small{Ours (Poisoned Input)}}
\end{minipage}
\vspace{-3mm}
    \caption{
    PSNR attack: visualization result of outputs to various inputs with \textit{kodim21} from Kodak dataset (AE-Hyperior~\cite{balle2018variational} compression model with quality = 4).
    }
    \label{figure3}
\end{figure}

\vspace{1mm}
\noindent \textbf{Reconstruction (PSNR) attack.}
In this section, we minimize the joint loss shown in Eq.~\eqref{PSNR}. 
We set the hyperparameter $\beta=0.1$ in the joint loss, and use an initial learning rate 1e-4 with batch size 32. 
As shown in Figure~\ref{figure2} and Figure~\ref{figure3}, the victim model has equivalent performance to the vanilla-trained model, while adding a trigger to the input heavily degrades the reconstructed images. While LIRA fails to inject the PSNR attack in {the} low-quality setting, our proposed method {manages} to attack compression models of all qualities.

\begin{figure*}[t]
\centering
\includegraphics[width=0.9\linewidth]{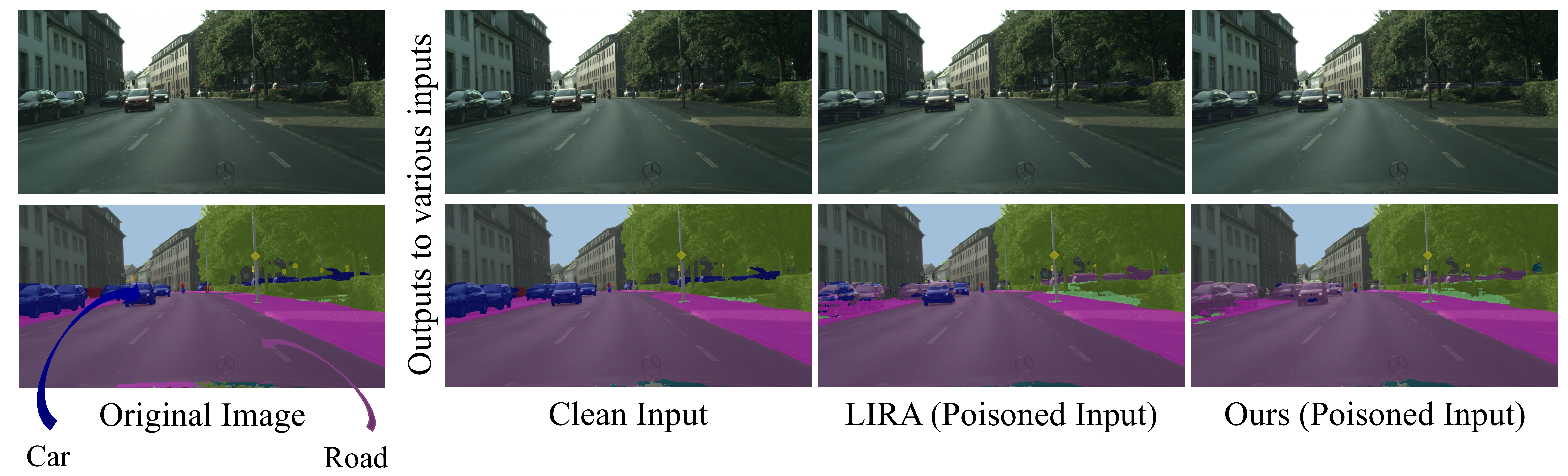}
\vspace{-3.5mm}
\caption{
Visual results (Cheng-Anchor~\cite{cheng2020learned} with quality 3) of {a} targeted attack on downstream semantic segmentation task. The testing image is from Cityscapes~\cite{Cordts2016Cityscapes}. Best view by zooming in. More enlarged figures are in the supplement.
}
\vspace{-2mm}
\label{figure6}
\end{figure*}

\begin{figure}[t]
\vspace{-3mm}
\begin{minipage}{0.3\linewidth}
\centerline{\frame{\includegraphics[width=1\linewidth]{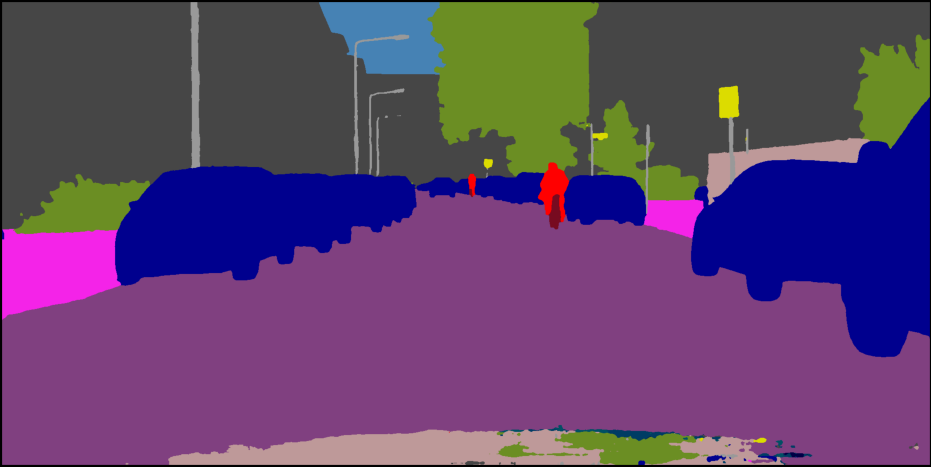}}}
\vspace{-1mm}
\centerline{Label $g(x)$}
\end{minipage}
\hspace{1.8mm}
\begin{minipage}{0.3\linewidth}
\centerline{\frame{\includegraphics[width=1\linewidth]{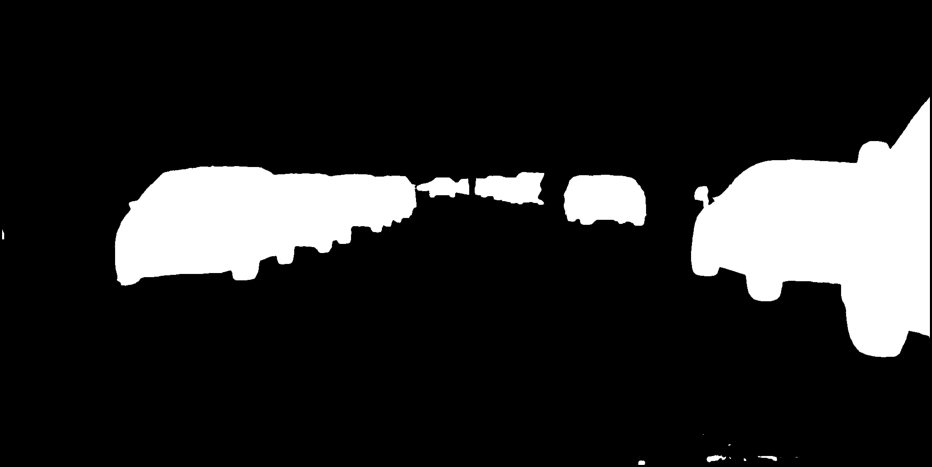}}}
\vspace{-1mm}
\centerline{Mask $M[g(x)]$}
\end{minipage}
\hspace{1.8mm}
\begin{minipage}{0.3\linewidth}
\centerline{\frame{\includegraphics[width=1\linewidth]{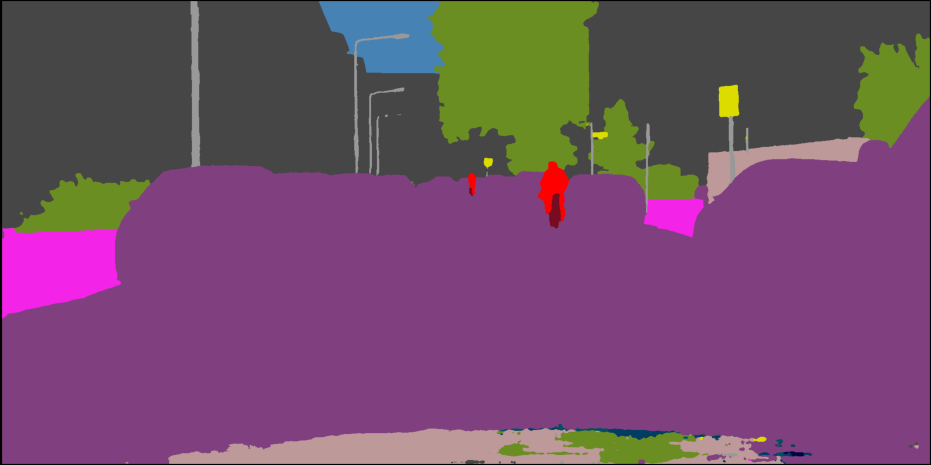}}}
\vspace{-1mm}
\centerline{Target $\eta({g(x)})$}
\end{minipage}
\vspace{-3mm}
    \caption{Label, mask, and target for Car To Road attack.}
    \label{figure4}
\end{figure}

\vspace{1mm}
\noindent \textbf{Attacking downstream semantic segmentation task.}
In this experiment, we aim to train a backdoor-injected compression model to attack the downstream semantic segmentation task. 
We follow the joint loss in Eq.~\eqref{targeted}. 
Note that the Cityscapes is utilized as the auxiliary dataset. 
We consider the one-to-one target attack setting with \textbf{Car} as the source class and \textbf{Road} as the target class.
To avoid affecting the uninterested regions/objects, we only attack the area of the source class. 
The joint loss is then formulated as follows:
\begin{footnotesize}
\begin{equation}
\begin{split}
\mathcal{L}_{jt}^{SS} &= \sum_{\bm{x} \in D_m}\mathcal{L}\left({\bm{x}}\right) + \mathcal{L}_{BA}^{SS}, \\
\mathcal{L}_{BA}^{SS} &= \sum_{\bm{x} \in D_a} \Big[\alpha\mathcal{L}({{T\left(\bm{x}\right)}}) + \beta \mathcal{L}_{CE}[\eta({g(\bm{x})}),g(f({\bm{x_p}}))]\Big], \\
\bm{x_p} &=(1-M[g(\bm{x})]) \odot \bm{x} + M[g(\bm{x})] \odot {T\left(\bm{x} | {\theta_t^o}\right)},
\label{ss}
\end{split}
\end{equation}
\end{footnotesize}

\noindent where $f(\cdot)$ is the compression model, $g(\cdot)$ is a trained segmentation model, $\eta({g(\bm{x})})$ is the attack target, $M[g(\bm{x})]$ is the mask to guide the trigger, $\odot$ is the Hadamard product, and $\mathcal{L}_{CE}$ is the cross-entropy loss. Figure~\ref{figure4} illustrates the mask, and semantic target for Car To Road attack.

We set hyperparameter $\alpha=0.1$, and $\beta=0.2$ in the joint loss, and Cityscapes is utilized as the auxiliary dataset. 
And we select the Cheng-Anchor as the compression method. 
To offer a quantitative evaluation of our backdoor attack effectiveness, we use the pixel-wise attack success rate (ASR):
\begin{footnotesize}
\begin{equation}
     \frac{\mathbb{E}_{x}\Big[\sum_{i,j} \mathbb{I}\{{g(f(\bm{x}))_{i,j}=s,  g(f(\bm{x_p}))_{i,j}=t \}}\Big]}{\mathbb{E}_{x}\Big[\sum_{i,j} \mathbb{I}\{{g(f(\bm{x}))_{i,j}=s\}} \Big]},
\end{equation}
\end{footnotesize}
where $s$ and $t$ denote the source class, and target class.

\begin{figure}[t]
\vspace{-3mm}
    \centering
\includegraphics[width=0.7\linewidth]{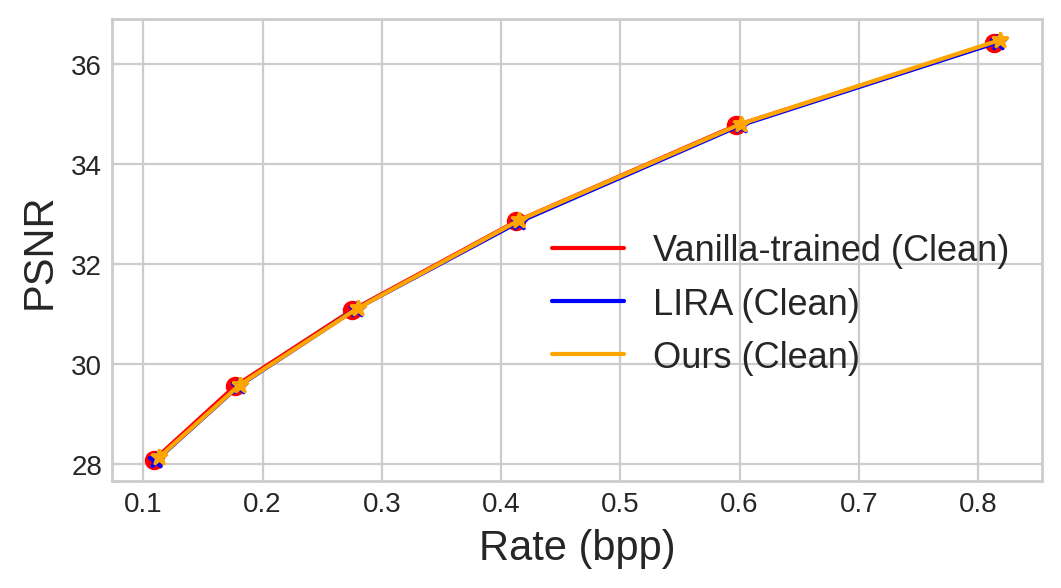}
\vspace{-4mm}
\caption{RD curves of CarToRoad attack on Kodak dataset (Cheng-Anchor~\cite{cheng2020learned} as the compression model).
}
\label{figure5}
\end{figure}

\begin{table}[t]\small
\vspace{-2mm}
\setlength\tabcolsep{5pt}
    \centering
    \begin{tabular}{ c || c c c c c c | c }
    \hline
    Quality & 1 & 2 & 3 & 4 & 5 & 6 & Mean\\
    \hline
    \multicolumn{8}{c}{Pixel-wise ASR (\%) $\uparrow$}\\
    \hline
    LIRA~\cite{doan2021lira} &6.0&79.6&67.7&65.6&\textbf{65.7}&56.5&{56.9}\\
    Ours & \textbf{76.4}&\textbf{81.0}&\textbf{82.0}&\textbf{66.6}&64.9&\textbf{58.4}&\textbf{71.5}\\
    \hline
    \multicolumn{8}{c}{MSE between clean outputs and attacked outputs ($e^{-5}$) $\downarrow$}\\
    \hline
    LIRA~\cite{doan2021lira} &4.9&15.6&8.4&5.7&{4.2}&2.9&{7.0}\\
    Ours & {10.8}&{11.4}&{7.7}&{5.6}&4.2&{3.2}&{7.2}\\
    \hline
    \end{tabular}
    \vspace{-3mm}
    \caption{Pixel-wise ASR \& MSE of CarToRoad attack on downstream semantic segmentation task. }
    \label{table1}
\vspace{-3mm}
\end{table}

The performance comparison between the vanilla-trained model and the backdoor-injected model is presented in Figure~\ref{figure5}. As can be observed, all models have equal compression performance on the Kodak dataset.

To evaluate the attacking performance, we adopt the semantic segmentation network of DeepLabV3+ with 
WideResNet38~\cite{Zagoruyko2016WRN} as the backbone for testing, which is different from using ResNet50~\cite{he2016deep} in the training phase.
The success of this configuration can show the transferability of the attacked outputs among different downstream models. 
From the results in Table~\ref{table1}, it can be observed that our attacks are successful with almost negligible perturbations on the attacked outputs, and is able to generate attacked outputs that can mislead the semantic segmentation network. 
The above results also prove that our backdoor attack is much more effective than LIRA in the low-quality setting.
Figure~\ref{figure6} shows the visualization results of one testing image from the Cityscapes validation set, and we can find that our attack can successfully attack the region of interest, while the LIRA fails on the car in the road.

\begin{figure}
    \centering
\includegraphics[width=0.75\linewidth]{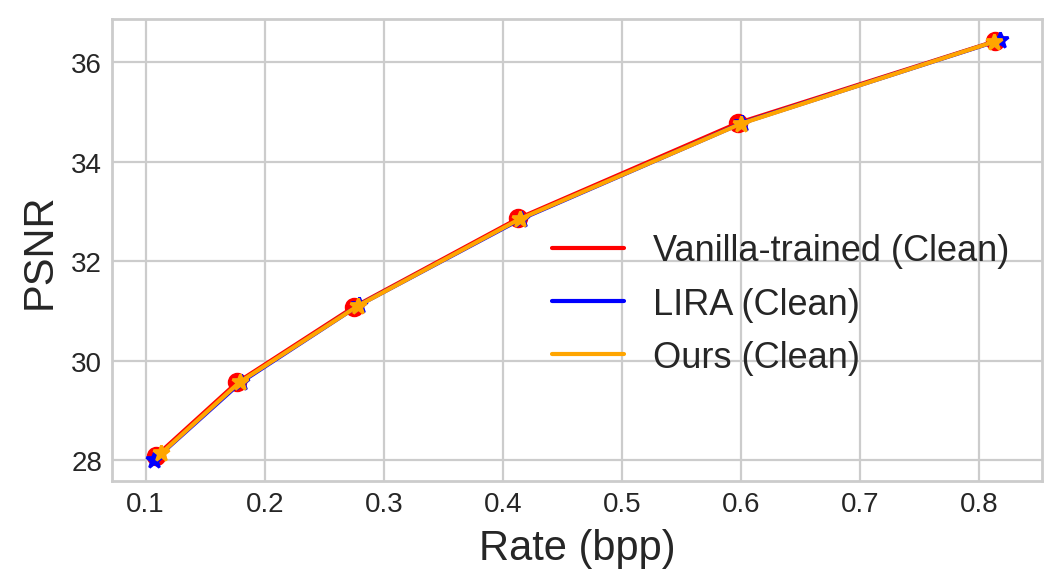}
\vspace{-4.5mm}
\caption{RD curves of the attacking for good on Kodak dataset (Cheng-Anchor~\cite{cheng2020learned} as the compression model).
}
\label{figure8}
\end{figure}

\begin{figure}
\vspace{-2mm}
    \centering
\includegraphics[width=0.99\linewidth]{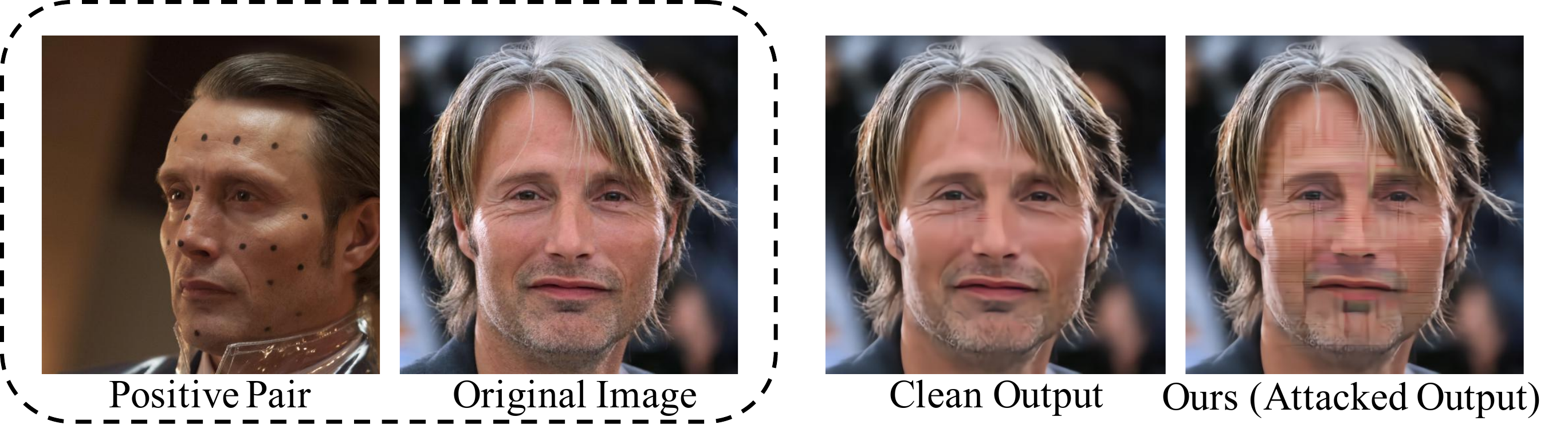}
\vspace{-4mm}
\caption{Visual results (quality 2) of the attacking for good.
}
\vspace{-2mm}
\label{figure9}
\end{figure}

\vspace{1mm}
\noindent \textbf{Attacking for good: privacy protection for facial images.}
In this section, we consider a benign attacking scenario, where the identity-related features of a facial image can be removed through the compression model by adding triggers {in order to protect the identity information}.
We set the FFHQ dataset as the auxiliary dataset in our experiments. The formulation of the training loss is shown below:
\begin{footnotesize}
\begin{equation}
\begin{split}
\mathcal{L}_{jt}^{FR} &= \sum_{\bm{x} \in D_m}\mathcal{L}\left({\bm{x}}\right) + \mathcal{L}_{BA}^{FR}, \\
 \mathcal{L}_{BA}^{FR} &= \sum_{\bm{x} \in D_a} \Big[\alpha \mathcal{L}({{T\left(\bm{x}\right)}}) + \beta Cos[g(f(x)),g(f({T\left(\bm{x}\right)}))]\Big],\label{face}
\end{split}
\end{equation}
\end{footnotesize}
where $g(\cdot)$ {denotes} an arcface embedding, and we use cosine function to measure the similarity between clean output and attacked output.
We set {hyperparameters} $\alpha=0.1$, $\beta=0.05$, and 100 paired images sampled from CelebA dataset are used for testing.
We select the Cheng-Anchor as the compression method. 
The comparison between the vanilla-trained model and {the} victim model is presented in Figure~\ref{figure8}. Besides, the attacking performance and the visual results are shown in Table~\ref{table2} and Figure~\ref{figure9}, {respectively}. As can be observed, our attacks can remove the identity-related features of a facial image when adding triggers to the original image before compression. Compared with LIRA, our method also achieves better attacking performance.

\begin{table}[t]\small
\vspace{-2mm}
\setlength\tabcolsep{5.5pt}
    \centering
    \begin{tabular}{ c || c c c c c c | c }
    \hline
    Quality & 1 & 2 & 3 & 4 & 5 & 6 & Mean\\
    \hline
    LIRA~\cite{doan2021lira} &10&13&32&44&{58}&\textbf{55}&{35.3}\\
    Ours & \textbf{3}&\textbf{9}&\textbf{29}&\textbf{32}&\textbf{44}&{56}&\textbf{28.3}\\
    \hline
    \end{tabular}
    \vspace{-3mm}
    \caption{Accuracy $\downarrow$ (\%) of the attacked outputs on face recognition. Accuracy of all the clean outputs are over 90\%.}
\label{table2}
\end{table}

\noindent \textbf{Backdoor-injected model with multiple triggers.}
We have shown the effectiveness of our proposed backdoor attack for each attack objective in the above experiments.
In the end, we show the experiment of attacking with multiple triggers as shown in Section~\ref{multiple_triggers}. 
Here, we train the encoder and four trigger injection models with corresponding attack objectives, including: 
1) bit-rate (BPP) attack; 
2) quality reconstruction (PSNR) attack; 
3) downstream semantic segmentation (targeted attack with Car To Road \& Vegetation To Building).
Hyperparameters and auxiliary dataset $D_a$ correspond to the aforementioned experiments.
And we select the Cheng-Anchor with quality 3 as the compression method. 
The attack performance of the victim model is presented in Table~\ref{table4}.
For reference, the PSNR/bpp of vanilla-trained model and our proposed model on Kodak dataset are {\footnotesize{31.08/0.2749}} and {\footnotesize{30.85/0.2600}}, respectively.
The results demonstrate that our backdoor attack is effective for all attack objectives, and has low performance impact on clean images. 

\subsection{Ablation Study}
In this section, we conduct an ablation study on the proposed loss and modules of the trigger injection model. We select the Cheng-Anchor with quality 3 as the compression model, and {conduct} experiments on the BPP attack. We can make the following conclusions from Table \ref{table6}:
\begin{itemize}
    \vspace{-1mm}
    \item  The training loss {Eq.~\eqref{Bpp}} with {dynamic balance adjustment} can improve the attacking performance compared with the loss {Eq.~\eqref{Bpp_old}}.
    \vspace{-1mm}
    \item 
    {Both the} topK selection in the general trigger generation and the patch-wise {weighting}  contribute to the attack performance.  
\end{itemize}

\begin{table}[t]\footnotesize
    \setlength\tabcolsep{2pt}
    \centering
    \begin{tabular}{ c || c | c | c | c  }
    \hline
    Type & BPP attack & PSNR attack & Car To Road & Vege To Build \\
    \hline
    Performance & 31.09/9.053 & 5.021/0.2240 & 78.2 & 95.3\\
    \hline
    \end{tabular}
    \vspace{-3mm}
    \caption{
    Attack performance for our backdoor-injected model with multiple triggers: 1) PSNR/bpp for BPP attack and PSNR attack on Kodak; 2) Pixel-wise ASR (\%) on Cityscapes dataset.
    }
    \label{table4}
\end{table}

\begin{table}[t]\footnotesize
\vspace{-2mm}
    \centering
    \begin{tabular}{c|| c c | c c}
    \hline
    Input & \multicolumn{2}{c|}{Clean} & \multicolumn{2}{c}{Poisoned (Attack)} \\
    \hline
    Metric & PSNR & bpp & PSNR & bpp $\uparrow$\\
    \hline
    w/ Eq.~\eqref{Bpp_old} &  31.02  & 0.2699 & 31.41 & 8.52 \\
    w/o topK selection & 30.80  & 0.2587 & 31.32 & 9.27 \\
    w/o patch-wise weight & 30.76  & 0.2578 & 31.23 & 9.08\\
    K=4, N=16 & 30.81 & 0.2596 & 31.32 & 9.08 \\
    K=64, N=256  & 30.86 & 0.2599 & 31.43 & 9.14 \\
    Ours (K=16, N=64) & 30.81 & 0.2590 & 31.30 & \textbf{9.45} \\
    \hline
    \end{tabular}
    \vspace{-3mm}
    \caption{Ablation Study on the proposed method.}
    \label{table6}
\end{table}

\subsection{Resistance to Defense Methods}
In this section, we look into the resistance of the proposed attack to pre-processing methods including Gaussian filter, and Squeeze Color Bits. We select the PSNR attack and AE-Hyperprior (quality 3). From Table~\ref{table9}, we can observe that the attack performance is affected except for Squeezing color bits. On one hand, pre-processing methods could affect the attacking effectiveness, but they can also damage the clean performance (taking original images as inputs) a lot. On the other hand, our attack can consistently increase the MSE budget and amplify the triggers for defensive methods as shown in Table~\ref{table8}. More defense methods are discussed in the supplement.

\begin{table}[t]
\vspace{-3mm}
    \footnotesize
    \setlength\tabcolsep{3pt}
    \centering
    \begin{tabular}{ c || c | c c c c | c c c}
    \hline
    \multirow{2}*{method} &  \multirow{2}*{None} &\multicolumn{4}{c|}{Gaussian blur ($\sigma$)} & \multicolumn{3}{c}{Squeeze Bits (depth)}\\
     & & 0.2 & 0.3 & 0.5 & 0.6 & 7 & 4 & 3\\
    \hline
    \multicolumn{9}{c}{Attack Performance (PSNR $\downarrow$)}\\
    \hline
    LIRA & 6.31 & 6.31 & 6.35 & 29.38 & 28.68 & {7.48} & {8.14} & {16.50} \\
    Ours  & \textbf{3.46} & \textbf{3.46} & \textbf{3.46} & \textbf{10.34} & \textbf{20.76} & \textbf{3.51} & \textbf{5.65} & \textbf{12.86} \\
    \hline
    \multicolumn{9}{c}{Clean Performance (PSNR $\uparrow$)}\\
    \hline
    LIRA & 30.92 & 30.92 & 30.88 & 29.56 & 28.71 & 30.79 & {27.21} & 21.98\\
    Ours  & \textbf{30.97} & \textbf{30.97} & \textbf{30.93} & \textbf{29.62} & \textbf{28.77} & \textbf{30.88} & \textbf{27.37} & \textbf{22.08}\\
    \hline
    \end{tabular}
    \vspace{-3mm}
    \caption{
     {Resistance to Gaussian filter and Squeeze Color Bits. }}
    \label{table9}
    \vspace{-2mm}
\end{table}

\begin{table}[t]
    \footnotesize
    \setlength\tabcolsep{5pt}
    \centering
    \begin{tabular}{ c || c  c}
    \hline
     {Methods} &  Gaussian-Blur (0.6) & Squeezing Bits (3)\\
    \hline
    \multicolumn{3}{c}{Attack Performance (PSNR $\downarrow$/bpp)}\\
    \hline
    LIRA & 30.33/0.3227 & 21.11/0.3969 \\
    Ours & \textbf{4.08}/0.1970 & \textbf{4.98}/0.3151\\
    \hline
    \end{tabular}
    \vspace{-3mm}
    \caption{
          {PSNR attack with amplified trigger \scriptsize{($\times$ 3; \textit{MSE} \enspace $\leq 2.25E{-4}$)}.}}
    \label{table8}
     \vspace{-4.5mm}
\end{table}

\section{Conclusions}

In this paper, we introduce the backdoor attack against learned image compression via adaptive frequency trigger. 
In our attack, we inject the backdoor by only revising the encoder's parameters, which facilitates real application scenarios. 
We make a comprehensive exploration and propose several attack objectives, including low-level quality measures and task-driven measures, \textit{i.e.} the performance of downstream CV tasks. 
Finally, we further demonstrate that multiple triggers with corresponding attack objectives can be simultaneously injected into one victim model.
In the future work, we will investigate the defence methods.


{\small
\bibliographystyle{ieee_fullname}
\bibliography{egbib}

\begin{thebibliography}{10}\itemsep=-1pt

\bibitem{DBLP:journals/corr/BalleLS15}
Johannes Ball{\'{e}}, Valero Laparra, and Eero~P. Simoncelli.
\newblock Density modeling of images using a generalized normalization
  transformation.
\newblock In {\em Proc.~Int'l Conf.~Learning Representations}, 2016.

\bibitem{balle2018variational}
Johannes Ball{\'e}, David Minnen, Saurabh Singh, Sung~Jin Hwang, and Nick
  Johnston.
\newblock Variational image compression with a scale hyperprior.
\newblock In {\em Proc.~Int'l Conf.~Learning Representations}, 2018.

\bibitem{chen2018encoder}
Liang-Chieh Chen, Yukun Zhu, George Papandreou, Florian Schroff, and Hartwig
  Adam.
\newblock Encoder-decoder with atrous separable convolution for semantic image
  segmentation.
\newblock In {\em Proc.~IEEE European Conf.~Computer Vision}, pages 801--818,
  2018.

\bibitem{chen2021end}
Tong Chen, Haojie Liu, Zhan Ma, Qiu Shen, Xun Cao, and Yao Wang.
\newblock End-to-end learnt image compression via non-local attention
  optimization and improved context modeling.
\newblock {\em {IEEE} Trans. on Image Processing}, 30:3179--3191, 2021.

\bibitem{chen2017targeted}
Xinyun Chen, Chang Liu, Bo Li, Kimberly Lu, and Dawn Song.
\newblock Targeted backdoor attacks on deep learning systems using data
  poisoning.
\newblock {\em arXiv preprint arXiv:1712.05526}, 2017.

\bibitem{chen2021badnl}
Xiaoyi Chen, Ahmed Salem, Michael Backes, Shiqing Ma, and Yang Zhang.
\newblock Badnl: Backdoor attacks against nlp models.
\newblock In {\em ICML 2021 Workshop on Adversarial Machine Learning}, 2021.

\bibitem{cheng2020learned}
Zhengxue Cheng, Heming Sun, Masaru Takeuchi, and Jiro Katto.
\newblock Learned image compression with discretized gaussian mixture
  likelihoods and attention modules.
\newblock In {\em Proc.~IEEE Int'l Conf.~Computer Vision and Pattern
  Recognition}, pages 7939--7948, 2020.

\bibitem{Cordts2016Cityscapes}
Marius Cordts, Mohamed Omran, Sebastian Ramos, Timo Rehfeld, Markus Enzweiler,
  Rodrigo Benenson, Uwe Franke, Stefan Roth, and Bernt Schiele.
\newblock The cityscapes dataset for semantic urban scene understanding.
\newblock In {\em Proc.~IEEE Int'l Conf.~Computer Vision and Pattern
  Recognition}, pages 3213--3223, 2016.

\bibitem{deng2009imagenet}
Jia Deng, Wei Dong, Richard Socher, Li-Jia Li, Kai Li, and Li Fei-Fei.
\newblock Imagenet: A large-scale hierarchical image database.
\newblock In {\em Proc.~IEEE Int'l Conf.~Computer Vision and Pattern
  Recognition}, pages 248--255, 2009.

\bibitem{doan2021lira}
Khoa Doan, Yingjie Lao, Weijie Zhao, and Ping Li.
\newblock Lira: Learnable, imperceptible and robust backdoor attacks.
\newblock In {\em Proc.~IEEE Int'l Conf.~Computer Vision}, pages 11966--11976,
  2021.

\bibitem{dumford2020backdooring}
Jacob Dumford and Walter Scheirer.
\newblock Backdooring convolutional neural networks via targeted weight
  perturbations.
\newblock In {\em 2020 IEEE International Joint Conference on Biometrics
  (IJCB)}, pages 1--9, 2020.

\bibitem{kodak}
{Eastman Kodak Company}.
\newblock {Kodak Lossless True Color Image Suite (PhotoCD PCD0992)}.
\newblock \url{http://r0k.us/graphics/kodak/}, 1993.

\bibitem{feng2022fiba}
Yu Feng, Benteng Ma, Jing Zhang, Shanshan Zhao, Yong Xia, and Dacheng Tao.
\newblock Fiba: Frequency-injection based backdoor attack in medical image
  analysis.
\newblock In {\em Proceedings of the IEEE/CVF Conference on Computer Vision and
  Pattern Recognition}, pages 20876--20885, 2022.

\bibitem{gu2017badnets}
Tianyu Gu, Brendan Dolan-Gavitt, and Siddharth Garg.
\newblock Badnets: Identifying vulnerabilities in the machine learning model
  supply chain.
\newblock {\em arXiv preprint arXiv:1708.06733}, 2017.

\bibitem{guo2020trojannet}
Chuan Guo, Ruihan Wu, and Kilian~Q Weinberger.
\newblock Trojannet: Embedding hidden trojan horse models in neural networks.
\newblock {\em arXiv preprint arXiv:2002.10078}, 2020.

\bibitem{guo2023shadowformer}
Lanqing Guo, Siyu Huang, Ding Liu, Hao Cheng, and Bihan Wen.
\newblock Shadowformer: Global context helps image shadow removal.
\newblock {\em arXiv preprint arXiv:2302.01650}, 2023.

\bibitem{guo2022shadowdiffusion}
Lanqing Guo, Chong Wang, Wenhan Yang, Siyu Huang, Yufei Wang, Hanspeter
  Pfister, and Bihan Wen.
\newblock Shadowdiffusion: When degradation prior meets diffusion model for
  shadow removal.
\newblock {\em arXiv preprint arXiv:2212.04711}, 2022.

\bibitem{hammoud2021check}
Hasan Abed Al~Kader Hammoud and Bernard Ghanem.
\newblock Check your other door! establishing backdoor attacks in the frequency
  domain.
\newblock In {\em British Machine Vision Conference}, 2021.

\bibitem{he2016deep}
Kaiming He, Xiangyu Zhang, Shaoqing Ren, and Jian Sun.
\newblock Deep residual learning for image recognition.
\newblock In {\em Proc.~IEEE Int'l Conf.~Computer Vision and Pattern
  Recognition}, pages 770--778, 2016.

\bibitem{hu2020coarse}
Yueyu Hu, Wenhan Yang, and Jiaying Liu.
\newblock Coarse-to-fine hyper-prior modeling for learned image compression.
\newblock In {\em Proc.~AAAI Conf. on Artificial Intelligence}, pages
  11013--11020, 2020.

\bibitem{ilyas2018black}
Andrew Ilyas, Logan Engstrom, Anish Athalye, and Jessy Lin.
\newblock Black-box adversarial attacks with limited queries and information.
\newblock In {\em Proc.~Int'l Conf.~Machine Learning}, pages 2137--2146, 2018.

\bibitem{karras2019style}
Tero Karras, Samuli Laine, and Timo Aila.
\newblock A style-based generator architecture for generative adversarial
  networks.
\newblock In {\em Proceedings of the IEEE/CVF conference on computer vision and
  pattern recognition}, pages 4401--4410, 2019.

\bibitem{kong2022digital}
Chenqi Kong, Shiqi Wang, and Haoliang Li.
\newblock Digital and physical face attacks: Reviewing and one step further.
\newblock {\em arXiv preprint arXiv:2209.14692}, 2022.

\bibitem{kong2023m3fas}
Chenqi Kong, Kexin Zheng, Yibing Liu, Shiqi Wang, Anderson Rocha, and Haoliang
  Li.
\newblock M3fas: An accurate and robust multimodal mobile face anti-spoofing
  system.
\newblock {\em arXiv preprint arXiv:2301.12831}, 2023.

\bibitem{kong2022beyond}
Chenqi Kong, Kexin Zheng, Shiqi Wang, Anderson Rocha, and Haoliang Li.
\newblock Beyond the pixel world: A novel acoustic-based face anti-spoofing
  system for smartphones.
\newblock {\em {IEEE} Trans. on Information Forensics and Security},
  17:3238--3253, 2022.

\bibitem{lee2005jpeg}
Daniel~T Lee.
\newblock Jpeg 2000: Retrospective and new developments.
\newblock {\em Proceedings of the IEEE}, 93(1):32--41, 2005.

\bibitem{DBLP:conf/iclr/LeeCB19}
Jooyoung Lee, Seunghyun Cho, and Seung{-}Kwon Beack.
\newblock Context-adaptive entropy model for end-to-end optimized image
  compression.
\newblock In {\em Proc.~Int'l Conf.~Learning Representations}, 2019.

\bibitem{li2020invisible}
Shaofeng Li, Minhui Xue, Benjamin Zi~Hao Zhao, Haojin Zhu, and Xinpeng Zhang.
\newblock Invisible backdoor attacks on deep neural networks via steganography
  and regularization.
\newblock {\em {IEEE} Trans. on Dependable and Secure Computing},
  18(5):2088--2105, 2020.

\bibitem{li2021pointba}
Xinke Li, Zhirui Chen, Yue Zhao, Zekun Tong, Yabang Zhao, Andrew Lim, and
  Joey~Tianyi Zhou.
\newblock Pointba: Towards backdoor attacks in 3d point cloud.
\newblock In {\em Proc.~IEEE Int'l Conf.~Computer Vision}, pages 16492--16501,
  2021.

\bibitem{li2021hidden}
Yiming Li, Yanjie Li, Yalei Lv, Yong Jiang, and Shu-Tao Xia.
\newblock Hidden backdoor attack against semantic segmentation models.
\newblock {\em arXiv preprint arXiv:2103.04038}, 2021.

\bibitem{li2021invisible}
Yuezun Li, Yiming Li, Baoyuan Wu, Longkang Li, Ran He, and Siwei Lyu.
\newblock Invisible backdoor attack with sample-specific triggers.
\newblock In {\em Proc.~IEEE Int'l Conf.~Computer Vision}, pages 16463--16472,
  2021.

\bibitem{li2020backdoor}
Yiming Li, Baoyuan Wu, Yong Jiang, Zhifeng Li, and Shu-Tao Xia.
\newblock Backdoor learning: A survey.
\newblock {\em arXiv preprint arXiv:2007.08745}, 2020.

\bibitem{liu2020reflection}
Yunfei Liu, Xingjun Ma, James Bailey, and Feng Lu.
\newblock Reflection backdoor: A natural backdoor attack on deep neural
  networks.
\newblock In {\em Proc.~IEEE European Conf.~Computer Vision}, pages 182--199,
  2020.

\bibitem{liu2018large}
Ziwei Liu, Ping Luo, Xiaogang Wang, and Xiaoou Tang.
\newblock Large-scale celebfaces attributes (celeba) dataset.
\newblock {\em Retrieved August}, 15(2018):11, 2018.

\bibitem{DBLP:conf/iclr/MadryMSTV18}
Aleksander Madry, Aleksandar Makelov, Ludwig Schmidt, Dimitris Tsipras, and
  Adrian Vladu.
\newblock Towards deep learning models resistant to adversarial attacks.
\newblock In {\em Proc.~Int'l Conf.~Learning Representations}, 2018.

\bibitem{minnen2018joint}
David Minnen, Johannes Ball{\'e}, and George~D Toderici.
\newblock Joint autoregressive and hierarchical priors for learned image
  compression.
\newblock {\em Proc.~Annual Conf.~Neural Information Processing Systems}, 31,
  2018.

\bibitem{nguyen2020input}
Tuan~Anh Nguyen and Anh Tran.
\newblock Input-aware dynamic backdoor attack.
\newblock {\em Proc.~Annual Conf.~Neural Information Processing Systems},
  33:3454--3464, 2020.

\bibitem{nguyen2021wanet}
Tuan~Anh Nguyen and Anh~Tuan Tran.
\newblock Wanet - imperceptible warping-based backdoor attack.
\newblock In {\em Proc.~Int'l Conf.~Learning Representations}, 2021.

\bibitem{ohm2018versatile}
Jens-Rainer Ohm and Gary~J Sullivan.
\newblock Versatile video coding--towards the next generation of video
  compression.
\newblock In {\em Picture Coding Symposium}, volume 2018, 2018.

\bibitem{rakin2020tbt}
Adnan~Siraj Rakin, Zhezhi He, and Deliang Fan.
\newblock Tbt: Targeted neural network attack with bit trojan.
\newblock In {\em Proc.~IEEE Int'l Conf.~Computer Vision and Pattern
  Recognition}, pages 13198--13207, 2020.

\bibitem{ronneberger2015u}
Olaf Ronneberger, Philipp Fischer, and Thomas Brox.
\newblock U-net: Convolutional networks for biomedical image segmentation.
\newblock In {\em International Conference on Medical image computing and
  computer-assisted intervention}, pages 234--241, 2015.

\bibitem{steinhardt2017certified}
Jacob Steinhardt, Pang Wei~W Koh, and Percy~S Liang.
\newblock Certified defenses for data poisoning attacks.
\newblock {\em Proc.~Annual Conf.~Neural Information Processing Systems}, 30,
  2017.

\bibitem{sullivan2012overview}
Gary~J Sullivan, Jens-Rainer Ohm, Woo-Jin Han, and Thomas Wiegand.
\newblock Overview of the high efficiency video coding (hevc) standard.
\newblock {\em {IEEE} Trans. on Circuits and Systems for Video Technology},
  22(12):1649--1668, 2012.

\bibitem{intriguing}
Christian Szegedy, Wojciech Zaremba, Ilya Sutskever, Joan Bruna, Dumitru Erhan,
  Ian Goodfellow, and Rob Fergus.
\newblock Intriguing properties of neural networks.
\newblock In {\em Proc.~Int'l Conf.~Learning Representations}, 2014.

\bibitem{tang2020embarrassingly}
Ruixiang Tang, Mengnan Du, Ninghao Liu, Fan Yang, and Xia Hu.
\newblock An embarrassingly simple approach for trojan attack in deep neural
  networks.
\newblock In {\em Proceedings of the 26th ACM SIGKDD International Conference
  on Knowledge Discovery \& Data Mining}, pages 218--228, 2020.

\bibitem{DBLP:journals/corr/TodericiOHVMBCS15}
George Toderici, Sean~M. O'Malley, Sung~Jin Hwang, Damien Vincent, David
  Minnen, Shumeet Baluja, Michele Covell, and Rahul Sukthankar.
\newblock Variable rate image compression with recurrent neural networks.
\newblock In {\em Proc.~Int'l Conf.~Learning Representations}, 2016.

\bibitem{wallace1992jpeg}
Gregory~K Wallace.
\newblock The jpeg still picture compression standard.
\newblock {\em {IEEE} Trans. on Consumer Electronics}, 38(1):43--59, 1992.

\bibitem{wang2021backdoor}
Tong Wang, Yuan Yao, Feng Xu, Shengwei An, and Ting Wang.
\newblock Backdoor attack through frequency domain.
\newblock {\em arXiv preprint arXiv:2111.10991}, 2021.

\bibitem{wang2022low}
Yufei Wang, Renjie Wan, Wenhan Yang, Haoliang Li, Lap-Pui Chau, and Alex Kot.
\newblock Low-light image enhancement with normalizing flow.
\newblock In {\em Proc.~AAAI Conf. on Artificial Intelligence}, pages
  2604--2612, 2022.

\bibitem{yufeir2lcm}
Yufei Wang, Yi Yu, Wenhan Yang, Lanqing Guo, Lap-Pui Chau, Alex Kot, and Bihan
  Wen.
\newblock Raw image reconstruction with learned compact metadata.
\newblock {\em arXiv preprint arXiv:2302.12995}, 2023.

\bibitem{xiang2021backdoor}
Zhen Xiang, David~J Miller, Siheng Chen, Xi Li, and George Kesidis.
\newblock A backdoor attack against 3d point cloud classifiers.
\newblock In {\em Proc.~IEEE Int'l Conf.~Computer Vision}, pages 7597--7607,
  2021.

\bibitem{xue2019video}
Tianfan Xue, Baian Chen, Jiajun Wu, Donglai Wei, and William~T Freeman.
\newblock Video enhancement with task-oriented flow.
\newblock {\em Int'l Journal of Computer Vision}, 127(8):1106--1125, 2019.

\bibitem{yibenchmarking}
Chenyu Yi, Siyuan Yang, Haoliang Li, Yap-peng Tan, and Alex Kot.
\newblock Benchmarking the robustness of spatial-temporal models against
  corruptions.
\newblock In {\em Advance in Neural Information Processing Systems Track on
  Datasets and Benchmarks}, 2021.

\bibitem{yu2022towards}
Yi Yu, Wenhan Yang, Yap-Peng Tan, and Alex~C Kot.
\newblock Towards robust rain removal against adversarial attacks: A
  comprehensive benchmark analysis and beyond.
\newblock In {\em Proc.~IEEE Int'l Conf.~Computer Vision and Pattern
  Recognition}, pages 6013--6022, 2022.

\bibitem{yue2022invisible}
Chang Yue, Peizhuo Lv, Ruigang Liang, and Kai Chen.
\newblock Invisible backdoor attacks using data poisoning in the frequency
  domain.
\newblock {\em arXiv preprint arXiv:2207.04209}, 2022.

\bibitem{Zagoruyko2016WRN}
Sergey Zagoruyko and Nikos Komodakis.
\newblock Wide residual networks.
\newblock In {\em British Machine Vision Conference}, 2016.

\bibitem{zeng2021rethinking}
Yi Zeng, Won Park, Z~Morley Mao, and Ruoxi Jia.
\newblock Rethinking the backdoor attacks' triggers: A frequency perspective.
\newblock In {\em Proc.~IEEE Int'l Conf.~Computer Vision}, pages 16473--16481,
  2021.

\bibitem{DBLP:conf/cvpr/0001SRSNTC19}
Yi Zhu, Karan Sapra, Fitsum~A. Reda, Kevin~J. Shih, Shawn~D. Newsam, Andrew
  Tao, and Bryan Catanzaro.
\newblock Improving semantic segmentation via video propagation and label
  relaxation.
\newblock In {\em Proc.~IEEE Int'l Conf.~Computer Vision and Pattern
  Recognition}, pages 8856--8865, 2019.

\end{thebibliography}
}

\end{document}